\documentclass[a4paper,fleqn]{cas-sc}

\usepackage[authoryear,longnamesfirst]{natbib}
\usepackage{siunitx}
\usepackage{xcolor}
\usepackage{soul}
\usepackage{subfig}
\usepackage{upgreek}

\def\tsc#1{\csdef{#1}{\textsc{\lowercase{#1}}\xspace}}
\tsc{WGM}
\tsc{QE}
\begin{document}
\let\WriteBookmarks\relax
\def\floatpagepagefraction{1}
\def\textpagefraction{.001}

\shorttitle{Automated Full-Body Dermatoscopic Imaging}    
\shortauthors{Franchi et al.}  
\title[mode=title]{Multi-Robot Scanner for Automated Full-Body Dermoscopic Imaging}  

% Author 1
\author[1]{Valerio Franchi}[orcid=0000-0002-9592-9618]
\cormark[1]
\ead{valerio.franchi@udg.edu}
\ead[url]{}
\credit{Methodology, Software, Validation, Formal analysis, Investigation, Data curation, Writing -- original draft, Visualization}
\affiliation[1]{organization={University of Girona},
            city={Girona},
            country={Spain}}

% Author 2
\author[1]{Rafael Garcia}[orcid=0000-0002-1681-6229]
\ead{rafael.garcia@udg.edu}
\ead[url]{}
\credit{Conceptualization, Methodology, Investigation, Resources, Writing -- review \& editing, Supervision, Project administration, Funding acquisition}

% Author 3
\author[1]{Nuno Gracias}[orcid=0000-0002-4675-9595]
\ead{ngracias@silver.udg.edu}
\ead[url]{}
\credit{Writing -- review \& editing, Supervision}

% Author 4
\author[2]{Ricard Campos}[orcid=0000-0003-4718-468X]
\ead{ricard.campos@coronis.es}
\ead[url]{}
\credit{Methodology, Software, Validation, Formal analysis, Investigation, Data curation, Writing -- review \& editing, Visualization}
\affiliation[2]{organization={Coronis Computing S.L.},
            city={Girona},
            country={Spain}}

% Author 5
\author[2]{Josep Quintana}[orcid=0000-0003-3577-2011]
\ead{josep.quintana@coronis.es}
\ead[url]{}
\credit{Investigation, Writing -- review \& editing}

% Author 6
\author[2]{Sandra González-Villà}[orcid=0000-0002-5393-1664]
\ead{sandra.gonzalez@coronis.es}
\ead[url]{}
\credit{Methodology, Validation, Formal analysis, Data curation, Visualization}

% Author 7
\author[3]{Mark Ventura}[orcid=0009-0001-7357-2742]
\ead{mark.ventura@optotune.com}
\ead[url]{}
\credit{Writing -- review \& editing}
\affiliation[3]{organization={Optotune Switzerland AG},
            city={Dietikon},
            country={Switzerland}}

% Author 8
\author[4]{Nuria Ferrera}[orcid=0009-0008-9803-3009]
\ead{ferrera@recerca.clinic.cat}
\ead[url]{}
\credit{Validation, Formal analysis, Investigation, Resources, Writing -- review \& editing, Visualization}
\affiliation[4]{organization={Hospital Clínic de Barcelona},
            city={Barcelona},
            country={Spain}}

% Author 9
\author[4]{Clément Lenoir}[orcid=0009-0004-5174-9932]
\ead{lenoir@recerca.clinic.cat}
\ead[url]{}
\credit{Investigation, Resources, Writing -- review \& editing}

% Author 10
\author[4]{Josep Malvehy}[orcid=0000-0002-6998-914X]
\ead{jmalvehy@clinic.cat}
\ead[url]{}
\credit{Resources, Writing -- review \& editing}

\cortext[1]{Corresponding author}


\begin{abstract}
This paper outlines the specifications and design approach used to construct a full body imaging scanner capable of capturing skin lesions at a dermatoscopic level using cameras mounted on the end-effectors of four UR10 manipulators. The system possesses a view-planning algorithm capable of appropriately selecting the best camera position to acquire images of moles, a high-level controller to allow the manipulators to work simultaneously and a collision-detector that halts the manipulators when they make contact with an object or a person. We evaluate the system through real-patient full-body scans, comparing acquired images against contact dermoscopy and an existing total-body photography system (Vectra) across clinically relevant lesion features, and quantify true optical resolving power using a USAF 1951 resolution target, yielding a smallest resolvable feature size of \SI{22.1}{\micro\meter} for our scanner compared to  \SI{8.8}{\micro\meter} for contact dermoscopy. Results show the scanner consistently outperforms Vectra across most clinically relevant features and achieves comparable performance to contact dermoscopy for the majority of features assessed. By acquiring dermatoscopic-quality images automatically and without contact, and without requiring a separate manual dermoscopic examination, the scanner closes part of the gap between total-body photography and handheld dermoscopy, suggesting potential for future integration into screening workflows.
\end{abstract}


\begin{highlights}
\item Four-cobot system automates full-body dermatoscopic image acquisition
\item View-planning algorithm selects optimal camera positions for skin lesions
\item USAF 1951 target quantifies true optical resolution: \SI{22.1}{\micro\meter} vs \SI{8.8}{\micro\meter} 
\item System matches or exceeds Vectra and dermoscopy across most lesion features
\item Adaptive positioning enables consistent imaging across diverse body shapes
\end{highlights}

% Keywords

\maketitle

% Main text

\section{Introduction}
\label{sec:introduction}
Skin cancer incidence and mortality rates have risen over the past 50 years \citep{SkinPiotr}, with melanoma among its most aggressive forms \citep{TBPWinkler} . Dermatologists perform full body skin examinations to follow the progress of suspicious pigmented skin lesions. They are carried out using a dermatoscope (a hand-held microscope combined with a flashlight), checking each pigmented skin lesion individually in search of typical melanoma signs. This procedure can be highly time-consuming, particularly in patients with atypical mole syndrome or a large number of naevi. For this reason, total body photography (TBP) was developed to speed-up this process and provide doctors with a detailed map of the patient’s skin that can be used to track lesions as they change over time.

Commercial TBP systems have significantly improved whole-body skin documentation by rapidly acquiring standardized images of the entire skin surface. However, despite their high spatial coverage, they do not acquire images at dermoscopic resolution. Consequently, lesions identified during TBP still require a separate manual examination with a handheld dermoscope, making the workflow time-consuming. Examples of commercially-available TBP scanners include the Vectra WB360 by Canfield \citep{Vectra}, the DermoScan X2 by DermoScan GmbH \citep{DermoScan} the LumoScanner TBP System by Lumo Imaging \citep{Lumo} and the ATBM master 4th Generation by FotoFinder Systems GmbH \citep{ATBM}. Among these, the VECTRA WB360 is one of the most widely adopted clinical systems It consists of a whole-body 3D imaging system capturing the entire surface of the skin at high resolution with a single capture. It is composed of $92$ cameras with cross-polarized illumination (to eliminate reflections), and it is able to capture simultaneous images of both the front and back of the patient. The patient stands between these two half-scanners and a trigger captures the images from all of the cameras at the same time. It is currently employed in multiple hospitals for lesion detection and tracking over time \citep{VectraNews}. It has also been utilised in multiple clinical trials \citep{EfficacyVectra, MelanomaVectra, AustraliaVectra, ValidationVectra}.     

Existing full-body imaging scanners such as Vectra, DermoScan, LumoScanner and ATBM acquire high-resolution images of the entire skin surface, but do not reach true dermatoscopic-level detail. Lesions of interest identified during a TBP session must still be examined individually with a handheld dermatoscope, a manual step that remains time-consuming and operator-dependent. Non-contact dermoscopy itself is an established technique \citep{NonContactFerrera} , but to date it has required a human operator to position, focus and capture each lesion image individually. No commercially-available full-body imaging system automates this acquisition process using robotic manipulation. Our scanner is designed to close this gap: it automates the acquisition of dermatoscopic-quality images across the full body using four coordinated collaborative robots, removing the need for a separate manual dermoscopic examination step.

The aim of this paper is to present the design and validation of this scanner: a full-body imaging platform composed of four collaborative robots (cobots) that acquire simultaneous, non-contact dermatoscopic-level images of the patient's skin. High-resolution cameras equipped with liquid lenses are mounted on each cobot’s end-effector. These  lenses, based on two immiscible fluids with different refractive indexes, enable rapid refocusing across the full range of the scan without moving parts. We evaluate the resulting system through automated full-body scans on clinical subjects, comparing the acquired images with contact dermoscopy and a commercial total-body photography system, together with objective optical resolution measurements.

The remainder of this paper is structured as follows. Section \ref{section:mechanical_design} presents an overview of the scanner’s mechanical design, summarizing each individual component. Sections \ref{section:liquid_lens} and \ref{section:vision_system} provide details of the liquid lens and the vision system. Sections \ref{section:computer_vision_system} and \ref{section:scanning_procedure} describe the modules of the computer vision system (consisting of a calibration process, mole detection and view planning), and the body scan procedure, while Section \ref{section:gui} details the scanner’s human operator interface (both virtual and physical). Section \ref{section:results} presents results from the full-body scan procedure, including a comparison of lesion feature visibility across imaging modalities and an objective optical resolution assessment. And finally, Section \ref{section:conclusion} presents our conclusions and directions for future work. 

\section{System Overview}

\subsection{Mechanical Design}
\label{section:mechanical_design}

The scanner (Fig. \ref{fig:scanner_inside_lab}) comprises a set of sub-assemblies or sub-systems. Each component and subsystem is carefully designed and assembled to ensure the scanner’s precision and effectiveness in a clinical setting. The mechanical design encompasses several key components detailed in the following subsections: the sub-systems that form the core functionality of the scanner, the patient specifications that define the operational parameters, the cobot assembly that provides the robotic manipulation capabilities, and the gripper assembly that houses the imaging equipment. Each of these elements works in coordination to deliver precise, safe, and effective full-body dermatoscopic imaging.

\subsubsection{Sub-systems}

\begin{figure*}[t]
\centering
\subfloat[\centering]{\includegraphics[height=2.25in]{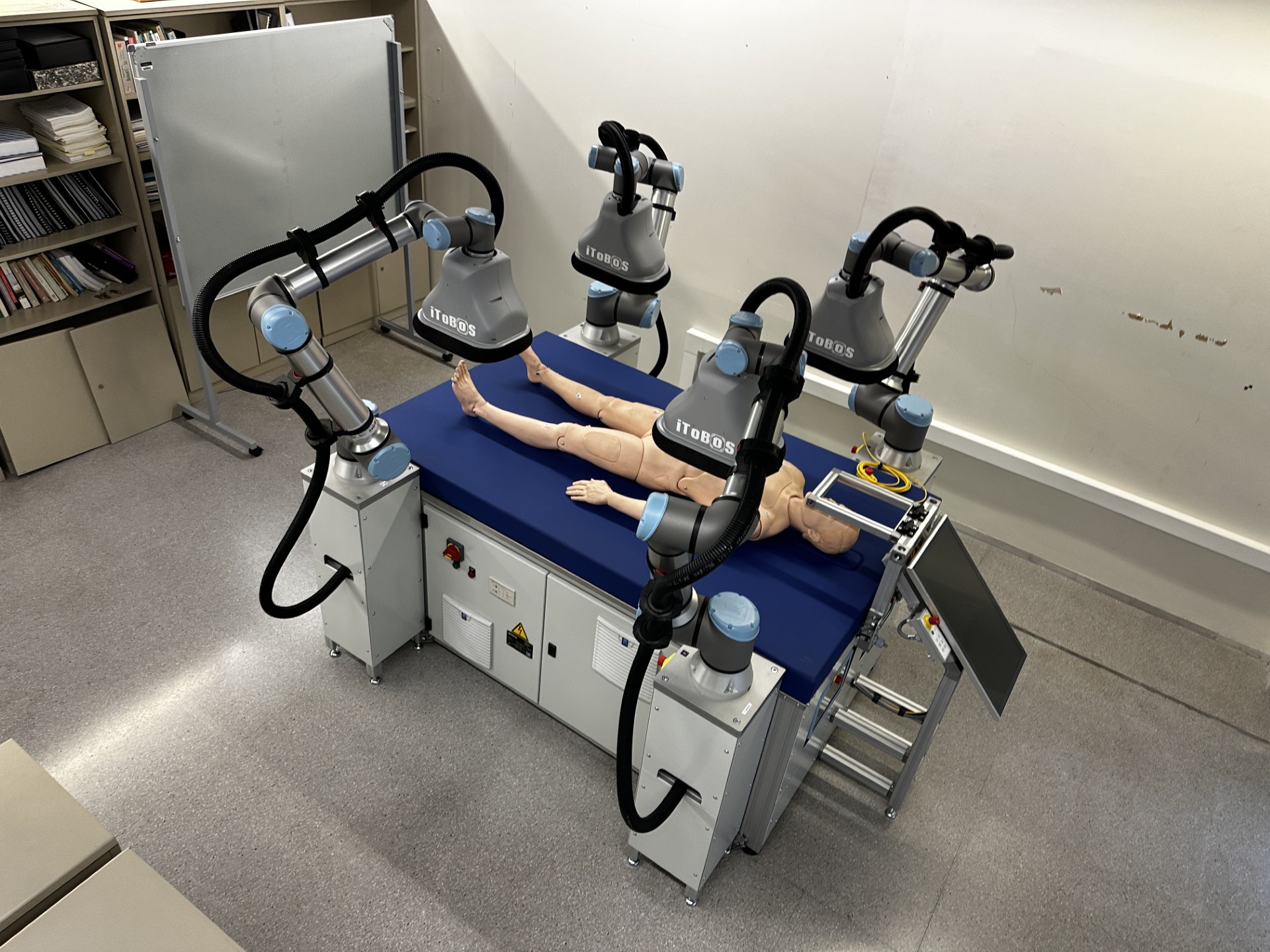}\label{fig:scanner_inside_lab}} 
\hspace{1cm}
\subfloat[\centering]{\includegraphics[height=2.25in]{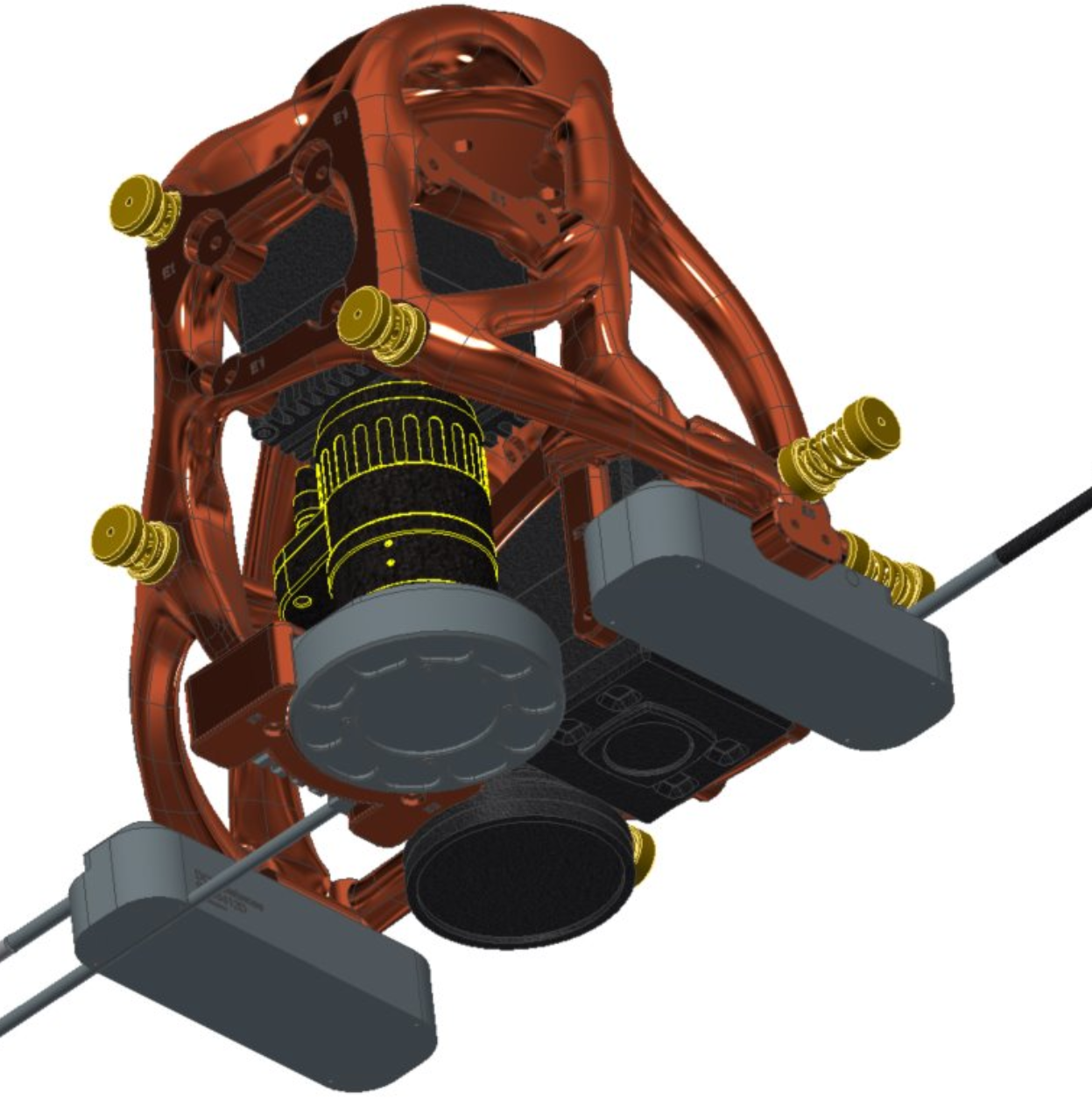}\label{fig:holder}}
\caption{(a) The scanner in the workshop with a mannequin for testing purposes. (b) CAD model of the gripper holder with the integrated hardware components (cameras, polarizers and ring-lights, represented in black and gray).}
\end{figure*}

\begin{itemize}
    \item The vision system (Section \ref{section:vision_system}) is composed of two groups of cameras. The first group of cameras acquires a set of images to perform a 3D reconstruction of the patient. Then any relevant pigmented skin lesions (larger than \SI{3}{\milli\meter} in diameter) are detected using the mole detector. Based on the 3D location of each lesion, the second group of cameras is placed in closer proximity to the lesion, at a distance of \SI{30}{\centi\meter}, and a dermoscopic image of the lesion is acquired using the dermoscopic camera. 
    \item The collaborative robots’ assembly with the cameras. The system consists of four robots, each of which covers an area of the patient’s body, which is divided into four quadrants. Each cobot’s end-effector houses a vision system, which is held by an additive manufactured holder (Fig. \ref{fig:holder}) and surrounded by a sphere-shaped protection. 
    \item A medical bed, which the patient lies on during the scanning process. 
    \item The main electrical cabinet that houses the core components of the scanner’s electrical system, including the Programmable Logic Controller (PLC) and all related control and electrical equipment. 
    % This centralised setup ensures efficient management and maintenance of the scanner’s electrical functions.
    \item  The central computer that serves as the operational hub for the scanner, equipped with the advanced vision algorithms essential for detecting the moles, generating the image stacks for dermoscopy and analysing the resulting images. The central computer is a custom build running Windows 10 with 256GB of RAM, an Intel Xeon Gold 6444Y processor (3.6GHz) and an NVIDIA RTX A4000 16GB GDDR6 graphics card.
    \item Human Machine Interface (HMI): the touchscreen interface operated by hospital personnel. 
\end{itemize}

\subsubsection{Patient specifications}

The dimensions of the systems are intended to be compatible with the following patient’s characteristics: 

\begin{itemize}
    \item Total height: \SIrange{140}{190}{\centi\meter}. 
    \item Thorax height (patient lying face up): \SIrange{20}{45}{\centi\meter}. 
    \item Elbow to elbow: \SIrange{40}{50}{\centi\meter}. 
    \item The patient is positioned lying down with their arms at the sides of the body, opening at an angle of around \ang{30} degrees, and the hands at approximately hip level. 
\end{itemize}

Informed consent was obtained for all the patients used in the experiments with the scanner.

\subsubsection{Cobot Assembly}

Collaborative robots (cobots) are designed to operate safely in human-shared environments. Following an evaluation of available market solutions, the UR10e CSI from Universal Robots  \citep{UR10e} was selected  for its advanced safety control system optimized for human-robot collaboration. This model features a \SI{12.5}{\kilo\gram} payload capacity and has a reach of \SI{130}{\centi\meter}. The assembly is composed of four cobots located around the patient. While performing the movements, the cobots can detect forces exerted at each point along their arms and will stop if the pressure sensed by the human is outside the acceptable limits. 

\subsubsection{Gripper Assembly}

The gripper-holder (Fig.  \ref{fig:holder}) was engineered  using organic Finite Element Analysis (FEA) methodologies. The design was optimized to significantly reduce weight, while enhancing critical mechanical properties, specifically reducing the moment of inertia and optimizing weight distribution for the intended application. This approach ensured a stable and controlled centre of mass for the complete system, thereby minimizing collision risks in collaborative human-robot environments. The design was then manufactured using Pa11 glass-filled material to achieve a lightweight and resistant polymer. The final gripper-holder protects the integrated vision hardware and minimizes the risk of collision from sharp edges.

\subsection{Liquid Lens System}
\label{section:liquid_lens}

A \textit{liquid lens} is an innovative camera technology that solves a common imaging problem: keeping objects at different distances in sharp focus. Traditional camera lenses have a fixed focus, meaning they can only capture clear images of objects within a specific distance range. Liquid lenses overcome this limitation by containing a special liquid that can change shape when electricity is applied, allowing the camera to instantly refocus from near to far objects without any moving parts.

For our scanner, we selected focus-tunable lenses that can adjust their focusing power across several diopters with remarkable precision and repeatability. This capability allows the system to capture a series of images of the same skin area, with each image focused at a different depth, ensuring that every detail is captured in sharp focus in at least one photograph. The lenses can change their focal power in just a few tens of milliseconds, making it possible to acquire all necessary images within minutes while keeping the patient comfortable and still throughout the process.

\begin{figure*}[t]
\centering
\subfloat[\centering]{\includegraphics[width=1.65in]{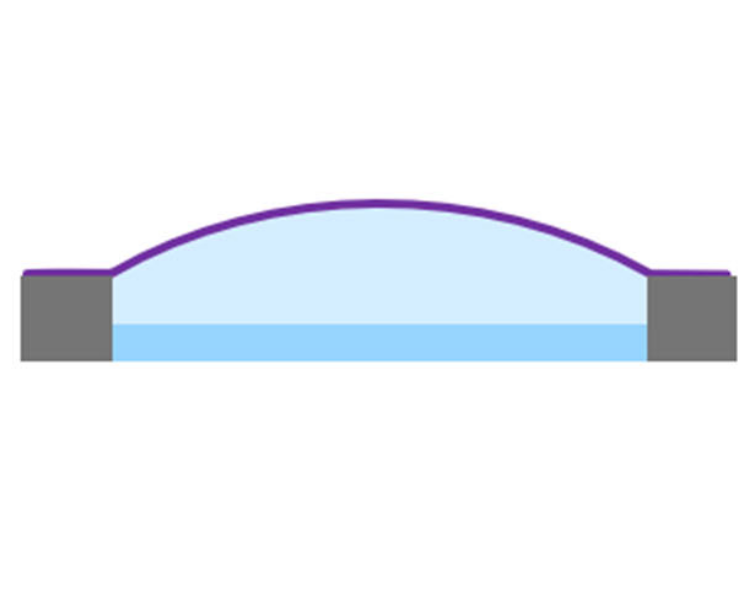}\label{fig:lens_horiz}} \hspace{0.5cm}
\subfloat[\centering]{\includegraphics[width=1.15in]{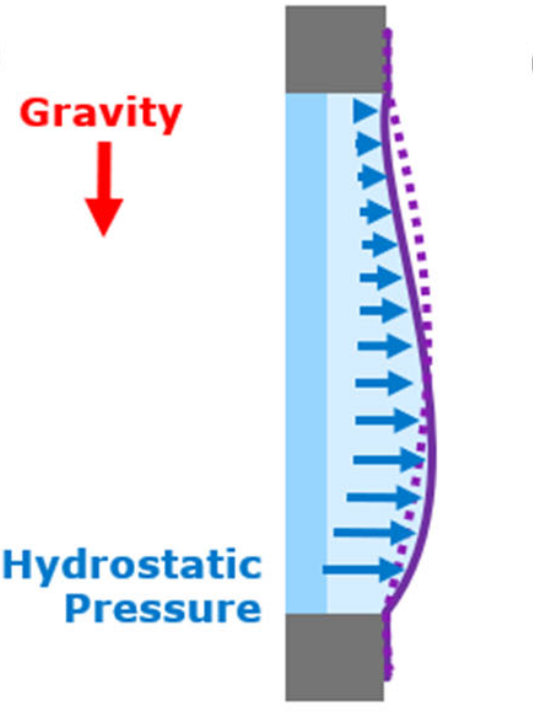}\label{fig:lens_vert}} \hspace{0.5cm}
\subfloat[\centering]{\includegraphics[width=1.15in]{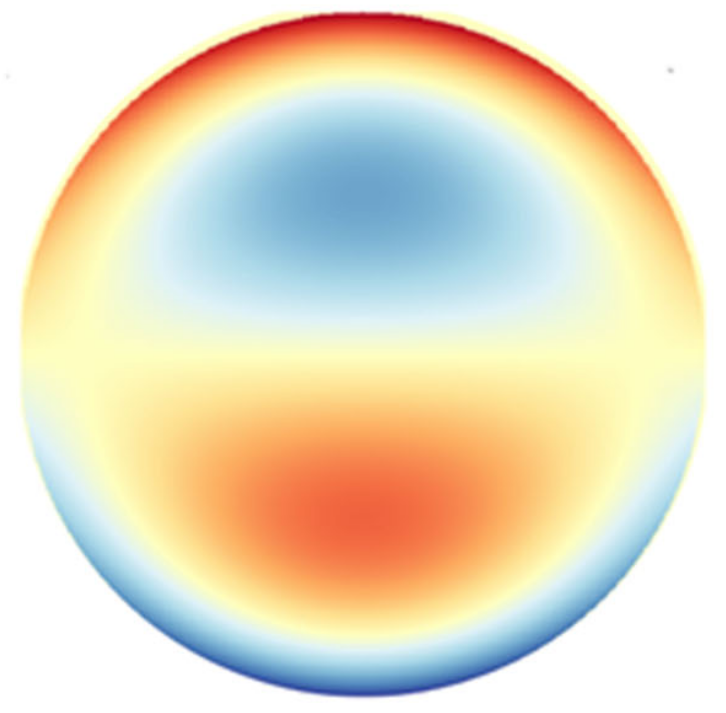}\label{fig:aberration}} \hspace{0.5cm}
\subfloat[\centering]{\includegraphics[width=1.8in]{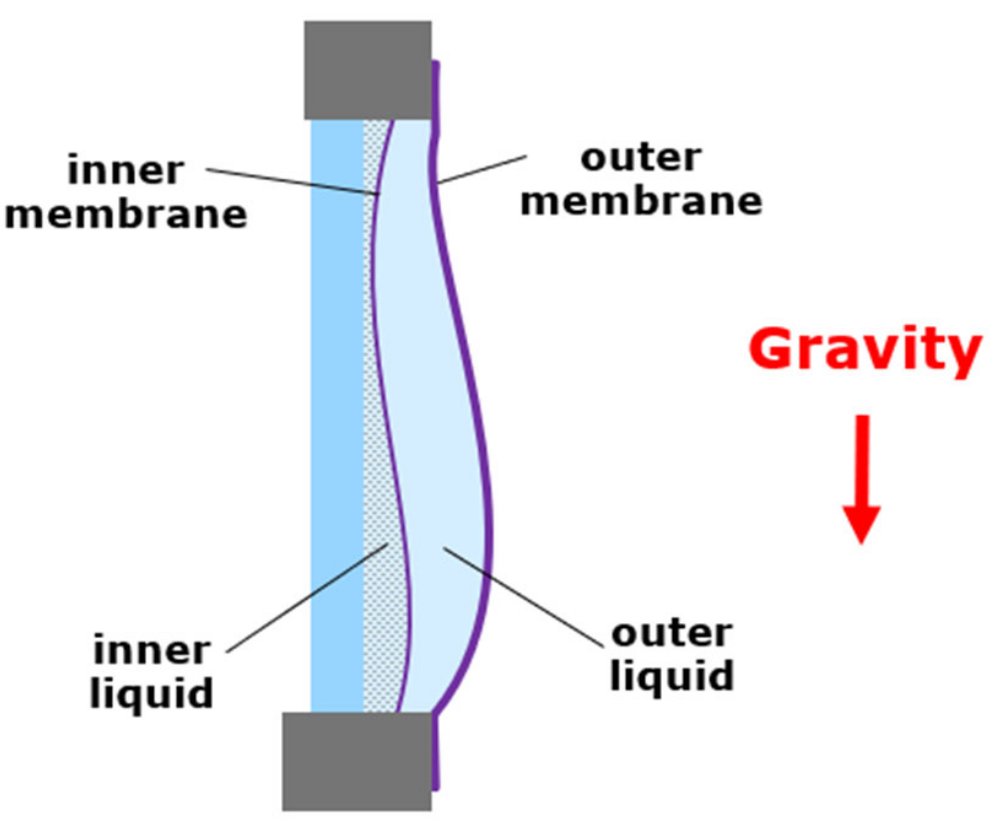}\label{fig:coma_compensation}} 
\caption{(a) Liquid lens in vertical (optical) axis configuration; (b) Liquid lens in horizontal axis configuration. The membrane deforms asymmetrically due to the effect of asymmetric hydrostatic pressure; (c) Optical power (arbitrary units) of a surface with coma aberration. The lower and higher optical power zones are represented in the bottom and top hemispheres respectively, mirroring the gravity-induced coma observed in liquid lenses; (d) In a coma-compensated lens, the shape taken by the second membrane combined with the right refractive indices of the two liquids can compensate for unwanted light diffraction.}
\label{fig:liquid_lens_coma}
\end{figure*}

While liquid lenses are the best solution for providing the fast and precise focusing needed, they do also have a limitation. Due to their liquid nature, performance is reduced when the optical axis is horizontal (Fig. \ref{fig:lens_horiz}) compared to when it is vertical (Fig. \ref{fig:lens_vert}). This is because, in the vertical axis configuration, the liquid pressure on the deformable membrane is evenly distributed across the entire lens surface, allowing the membrane to deform isotropically without developing any unwanted asymmetrical shape. However, when the lens is tilted - particularly with the optical axis in a horizontal position - the liquid’s weight causes the membrane to bulge outwards at the bottom due to hydrostatic pressure, while a corresponding void forms at the top, pulling the membrane inwards. This asymmetrical deformation results in an optical aberration equivalent to coma (Fig. \ref{fig:aberration}).

A solution was developed for this effect by incorporating a second liquid reservoir and a second membrane (Fig. \ref{fig:coma_compensation}). The weight of the heavier liquid deforms one membrane into a coma-like shape, while simultaneously displacing the lighter liquid upwards producing an \textit{inverted coma} shape on the second membrane. Choosing carefully the density and refractive index of the two liquids and the stiffness of the two membranes it is possible to passively compensate for the effect of gravity. Not all optical liquids and polymers are compatible - particularly under varying temperature and humidity conditions such as those expected for the full-body scanner (e.g. \SI{20}{\degree C}$\pm$\SI{5}{\degree C} in hospital environments) and during transport or storage in uncontrolled settings. To effectively reduce coma aberration, the properties of the two optical liquids (density and refractive index) and the two membranes (stiffness) must be properly matched. After rigorous testing, proprietary liquids OL1224 and OL1126 and membrane material SE1501 were selected for our particular case. 

This gravity compensation technology was integrated into the electrically tunable lens. This lens was selected for the full-body scanner’s optical system to provide the required tuning range while accommodating a large high-resolution $1$”-format sensor. The large sensor format was chosen to minimize the number of images needed to capture the patient’s full body.

\begin{figure*}[t]
\centering
\subfloat[\centering]{\includegraphics[height=2.15in]{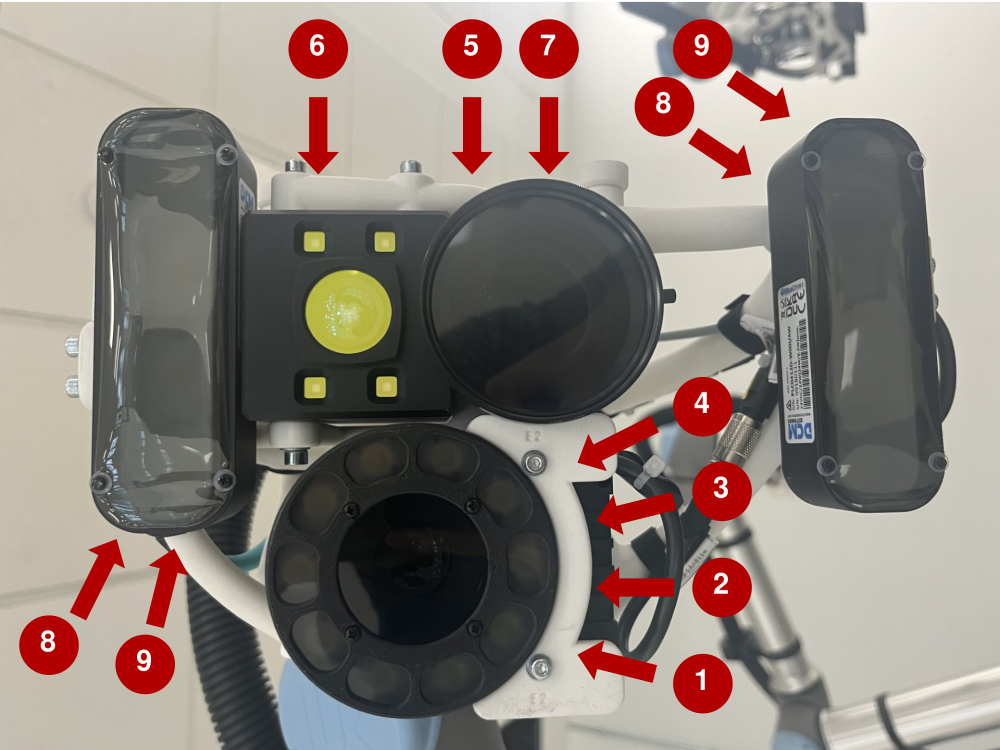}} \hspace{0.5cm}
\subfloat[\centering]{\includegraphics[height=2.15in]{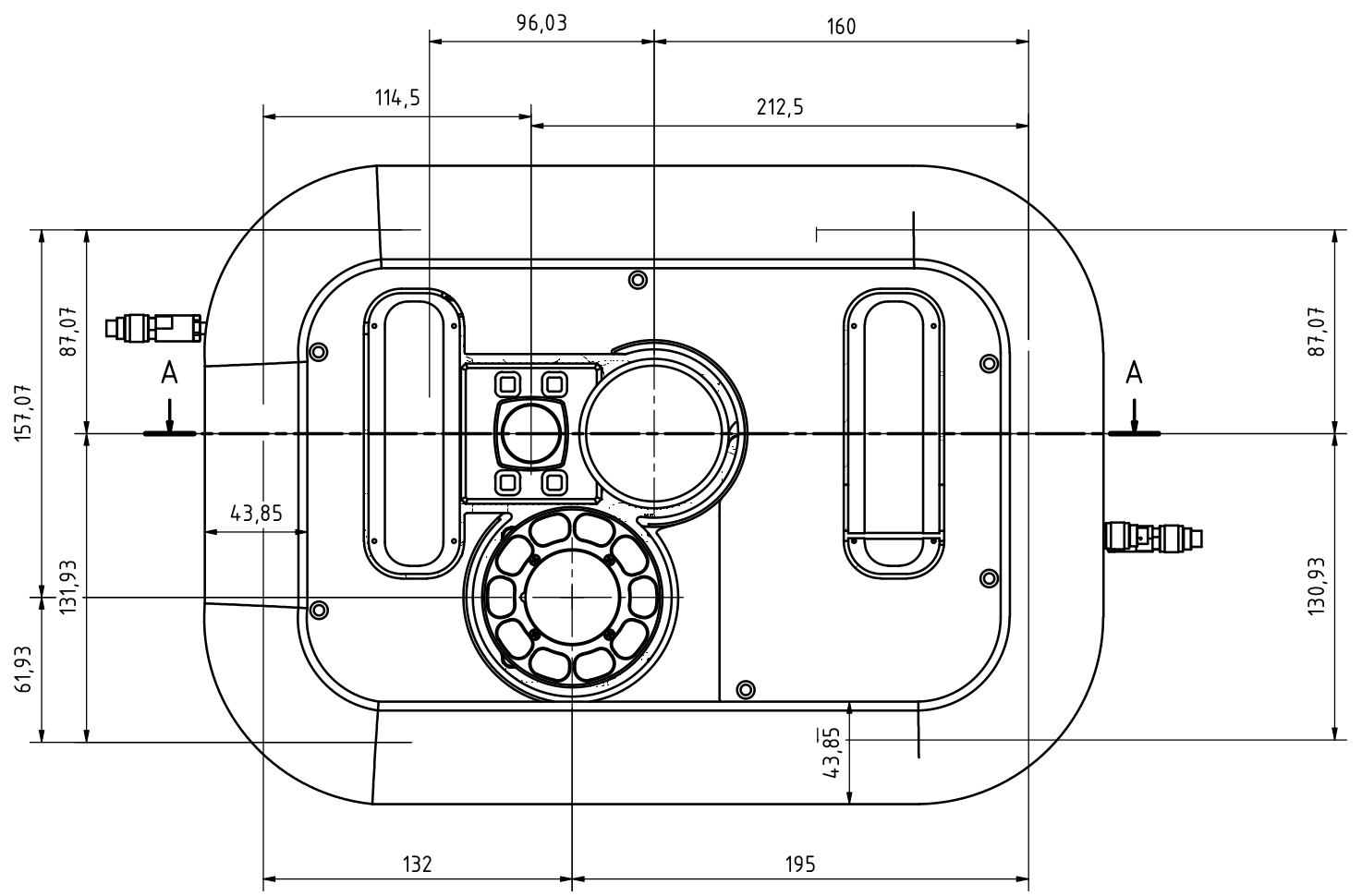}} 
\caption{(a) Image of the vision system mounted on the cobot’s end-effector without the protective housing. The numbers corresponding to those introduced at the start of Section \ref{section:vision_system}. (b) Schematic bottom-view of the end-effector showing camera positions and dimensions (in millimetres).}
\label{fig:vision_system}
\end{figure*} 

\subsection{Vision System}
\label{section:vision_system}

The imaging capabilities of the scanner rely on a sophisticated array of cameras and sensors working together to capture both the patient’s overall body structure and detailed close-up images of individual skin lesions. This multi-camera system combines different types of imaging technology: from wide-angle cameras that map the entire body to specialized dermoscopic cameras that can examine moles at microscopic detail. Each camera serves a specific purpose in the scanning process, and together they enable the system to automatically locate, approach and photograph skin lesions.

The main components of the vision system, illustrated in Fig. \ref{fig:vision_system}, are the following: 

\begin{enumerate}
    \item \textit{Oryx} 10GigE RGB Camera: dermoscopic camera.
    \item \textit{Optotune} EL-16-40-GTC Liquid Lens based on Evetar's \SI{75}{\milli\meter} lens:  dermoscopic lens.
    \item PR032-62 Linear Polarizer Filter: dermoscopic camera polarizer.
    \item \textit{ALB} Direct high-powered ring-light: dermoscopic cross polarized lighting.
    \item \textit{Lucid Triton} TRI200S-CC RGB Camera: 2D camera for mole detection.
    \item \textit{Lucid Helios2+} ToF IP67 3D Camera: 3D camera for patient reconstruction.
    \item Wide-Angle \SI{24}{MP} lens 1.1" \SI{8.5}{\milli\meter}/F2.5 Manual Iris.
    \item Polarizing Filter for the 2D camera
    \item PLC0412D Linear Light Projector: lighting system for the 2D camera.
\end{enumerate}

The dermoscopic camera is an \textit{Oryx} 10GigE RGB camera. The objective lenses are a custom design by Evetar Xiamen Leading Optics Co., Ltd \citep{EvetarOptics}, which integrates a liquid lens from Optotune to allow for a rapid change of focal length (detailed in Section \ref{section:liquid_lens} above). 

The 2D camera is a \textit{Triton} TRI200S-CC with LM8FC24M lenses from Kowa. The 3D camera is a \textit{Helios2+} Time of Flight (ToF). They are jointly employed to generate the patient’s 3D reconstruction, allowing for a proper mapping of the patient’s surface.

Additionally, the scanner’s lighting system has been specifically designed to complement the imaging process. Each pair of devices (the 2D and 3D cameras) are illuminated with two compact linear lights. These lights employ cross-polarization to eliminate potential skin reflections that could interfere with image quality. The integration of lighting  enhances the clarity and detail of the images captured, which is essential for accurate visualization of lesion features.

\subsection{Computer Vision System}
\label{section:computer_vision_system}

The computer vision system is composed of a calibration module that performs the initial camera calibrations on setup, and three operational modules (view-planning, mole detection and image enhancement) that work together to enable the mechatronic system to capture images of the patient’s skin lesions, perform mole detection and enhance the dermoscopic images for expert visual assessment of lesion features. Together, these modules automate the acquisition process end-to-end, removing the need for manual positioning, focusing, or image capture by a human operator.

\subsubsection{View-planning}
\label{sec:view_planning}

The view-planning module (described in detail in our previous work \citep{ViewPlanningFranchi}) utilizes the patient’s 3D reconstruction to perform two distinct tasks: to plan a full-body exploration at \SI{40}{\centi\meter} to capture 2D images that will be used for mole detection and to create more accurate higher-resolution 3D models for all sections of the patient’s body. By using both mole detections and finer-resolution reconstructions, the planner computes the optimal camera pose to capture each skin lesion at a distance of \SI{30}{\centi\meter}. It also takes into account a predefined safety distance to keep the patient’s discomfort to a minimum. 

\subsubsection{Mole detection}
\label{sec:mole_detection}

Mole detection is performed on the 2D images from the Triton camera using the YOLOv8 \citep{YoloVarghese} (You Only Look Once) architecture. The selection of the YOLO model family over other object detection models, such as RCNN and SSD, is based on several key considerations. The YOLO architecture is well-known for its real-time object detection capabilities, making it particularly well-suited for fast and efficient lesion detection in clinical environments. As skin lesion detection requires timely and accurate assessments, YOLO’s ability to achieve real-time inference aligns well with this demand. Moreover, YOLO’s unified detection approach, encompassing simultaneous object localization and classification, contributes to precise and accurate lesion detection. This integration minimizes localization errors and improves the efficient identification of skin lesions, further reinforcing its suitability for the task.

\subsubsection{Image enhancement}
\label{sec:image_enhancement}

Dermoscopic images captured by the Oryx camera equipped with our liquid lens are processed to provide detailed visual information on skin lesions. The liquid lens technology described in Section \ref{section:liquid_lens} enables a natural solution to a fundamental challenge in non-contact dermoscopy. As explained earlier, our liquid lenses can rapidly adjust their focal power in milliseconds, allowing us to capture multiple images of the same skin area with each image focused at a different depth. This capability becomes essential when surface topography exceeds any single depth of field, which would otherwise cause parts of the skin to appear out of focus. The system captures a stack of images at varying focal distances, then combines them through focus stacking techniques \citep{BuadesImageDenoising, EvangelidisImageAlignment, KanjarImageSharpness, SongImageFusion} to generate a single, focused dermoscopic image with complete depth coverage across all skin topographies. Additionally, any misalignments between images in the stack are corrected through precise sub-pixel registration. While this approach produces high-resolution, hyperfocused images, the resolution of individual lesions remains somewhat lower than traditional contact dermoscopy. To address this limitation, deep learning-based super-resolution techniques \citep{Yang2021GPEN, RealEsgran} are applied to enhance image detail and reveal additional structural information relevant to lesion feature visibility.

\begin{figure*}[t]
    \centering
    \includegraphics[width=0.3\textwidth]{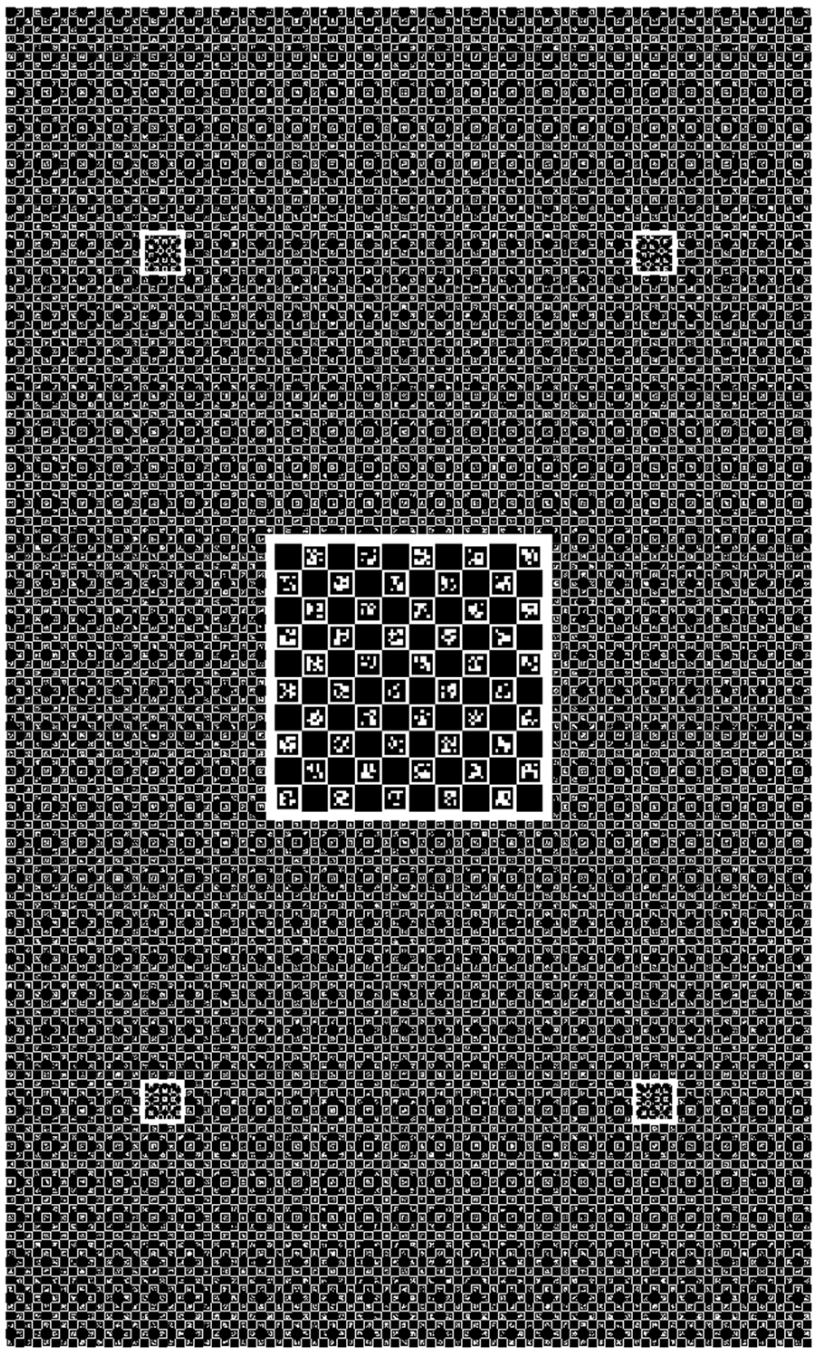}
    \caption{Calibration pattern with ChAruco patterns of \SI{1}{\centi\meter}, \SI{3}{\centi\meter} and \SI{0.5}{\centi\meter} squares designed specifically for the Triton, Helios and Oryx cameras, respectively.}
    \label{fig:calib_board}
\end{figure*}
\subsection{Calibration}
\label{section:calib}

Accurate 3D patient reconstruction and precise spatial coordination between the four cobots require careful estimation of several critical parameters. The intrinsic parameters of each camera (Triton, Helios, and Oryx) define the 3D-to-2D projection characteristics, following the standard pinhole model with radial and tangential distortion corrections \citep{FryerDistortion}. The Helios camera comes pre-calibrated by the manufacturer, with
parameters accessible through the provided SDK. Additionally, the system must determine the 3D pose (rotation and translation) of each camera relative to its cobot’s end-effector, as well as the base position of each robot relative to a global reference frame, enabling coordinated movement within a unified coordinate system.

The calibration process employs a specialized pattern affixed beneath the scanner table’s removable mattress. This pattern combines ChAruco markers  \citep{JuradoAruco}  of varying sizes adapting to each of the camera's field of view and resolution: large 1 cm squares covering most areas for the Triton camera, 3 cm squares in the center for the Helios camera, and small 0.5 cm squares in each quadrant for the Oryx cameras positioned near each robot base.

Calibration is performed by observing the pattern from multiple camera poses at each of the following sequential steps: 

\begin{enumerate}
    \item Intrinsic calibration of the Triton camera: each robot is guided through a series of pre-defined positions to capture images of the calibration pattern. Using standard non-linear optimization methods, the parameters of the pinhole and lens distortion models are computed for the Triton camera. 
    \item Eye-in-hand calibration of the Triton camera: the images and base-to-end-effector transformation (retrieved from the cobots) collected in Step 1 are used to compute the world-to-base and end-effector-to-triton transformations. 
    \item Stereo calibration between the Triton and Helios cameras: with the Triton camera being the reference frame within the end effector, the robot is guided through another series of pre-defined positions. These allow simultaneous image acquisition to capture the \SI{3}{\centi\meter}-per-square calibration pattern by the Helios camera and the \SI{1}{\centi\meter}-per-square pattern by the Triton camera. This enables computation of the relative transformation between the Triton and Helios frames, a necessary step for translating the 3D measurements from the Helios camera to the 2D image of the Triton camera. Using the parameters determined in Step 2, these 3D measurements can then be positioned within the global scanner frame. This capability is essential for real-time patient reconstruction. 
    \item Intrinsic calibration of the Oryx camera: following the same procedure as in Step 1, the robot is guided to a series of pre-defined positions to capture images of the small \SI{0.5}{\centi\meter}-per-square pattern near each robot’s base. Although this camera  does not follow a pinhole model due to the liquid lens, we assumed it nonetheless   and checked in practice that this assumption holds for our purposes.
    \item Stereo calibration between the Triton and Oryx cameras: as in Step 3, this step estimates the transformation between the Triton and Oryx cameras. The robot is again moved predefined positions, capturing images from both cameras, each observing at least part of its corresponding sub-pattern within the calibration pattern. After this step, all camera frames and their spatial relationships to the robots are known, thereby fully defining the scanner’s geometry.
    \item Liquid lens calibration: liquid lenses enable the selection of a specific dioptre value for each image. Since the patient’s 3D surface is reconstructed during scanning, it is essential to define the correct focus distance for each shot. This requires calibrating the relationship between dioptres and physical distances. To achieve this, the robot positions the Oryx camera orthogonally to the calibration pattern and triggers image capture at every available dioptre setting. The sharpness of each image is evaluated using the Normalised Gray Level Variance (NGLV) measure \citep{Santos1997EvaluationOA}, and the dioptre yielding the best in-focus result is recorded. This procedure is repeated at several known distances from the pattern, resulting in a set of optimal dioptre-distance pairs. A polynomial is then fitted to this dataset, enabling the system to determine the appropriate dioptre for any given distance during scanning. 

\end{enumerate}

\subsection{Scanning Procedure}
\label{section:scanning_procedure}

Once the patient is ready for scanning, the PLC checks that all the hardware and software components are ready. The device is then set to automatic mode, allowing the scanning procedure to begin. As a first step, the four robots move to a set of pre-defined global exploration poses to acquire an initial low-resolution 3D reconstruction of the patient using the Triton (2D, wide angle view) and Helios (3D) cameras. This initial textured 3D scan provides a first approximation of the patient’s skin surface geometry. Based on this map, the view planner (Section \ref{sec:view_planning}) computes a new set of camera positions to capture the skin at a resolution sufficient for the machine learning (ML) algorithms to detect skin lesions (Section \ref{sec:mole_detection}). 

Once the moles are detected in the 2D image, the 3D camera retrieves their precise coordinates in 3D space and computes their normal vectors, which characterize the local skin geometry around each lesion. Next, the planner computes a new, closer camera pose for each detected mole (Section \ref{sec:view_planning}). It then determines the optimal image acquisition sequence, thus commanding each robot to visit a different mole. At each target location, the system takes a picture using the Oryx camera.

Using the 3D model generated in the previous step—or a new 3D camera capture if the model is noisy in that region—the system estimates the depth of the patient’s skin relative to the Oryx camera.

This information, combined with the calibrated polynomial described in Section \ref{section:calib}, enables the system to define the minimum and maximum focus distances for image acquisition and adjust the liquid lens accordingly. An image is captured for each lens setting, producing a stack of images focused at varying depths. These are then combined into a single sharp image through focus stacking (Section \ref{sec:image_enhancement}). This process is repeated for each detected mole until all have been captured by the Oryx camera.

Once the scan is finished for one of the sides of the patient (i.e., forward- or backward facing), the PLC informs the operator to request the patient to turn  and performs the same sequence described above. 

\begin{figure*}[t]
\centering
\subfloat[\centering]{\includegraphics[width=2.5in]{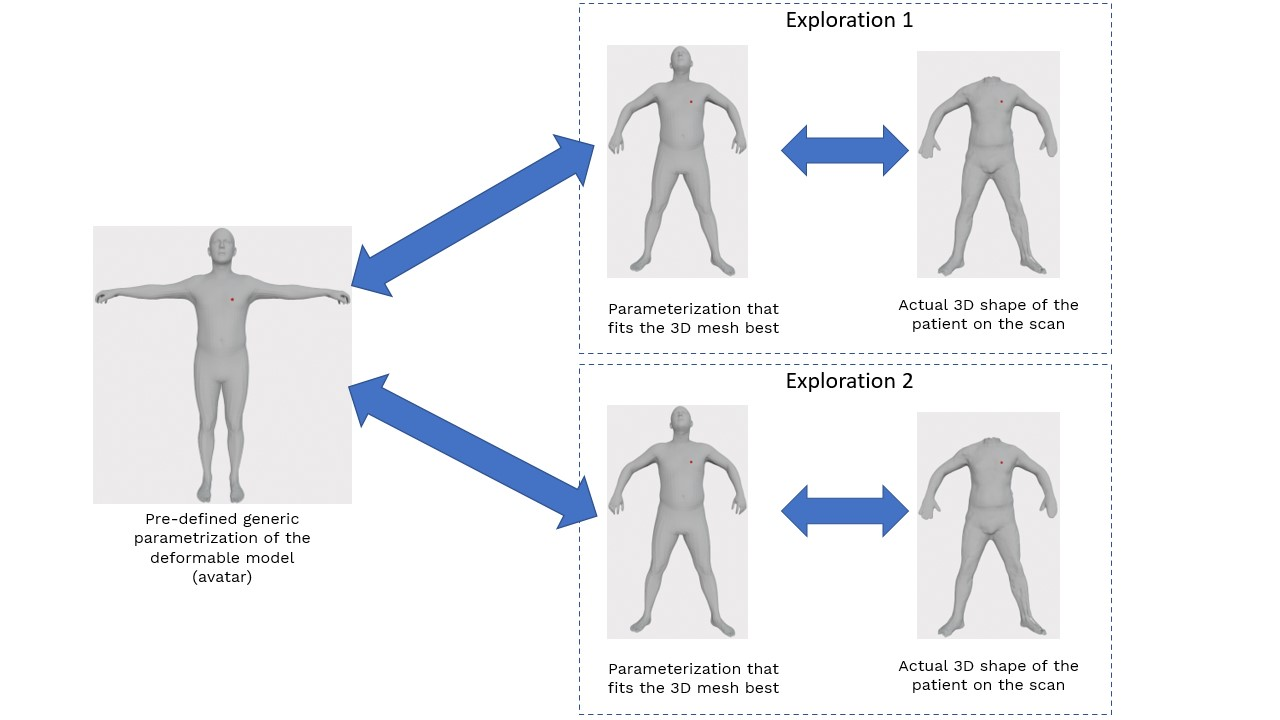}\label{fig:avatar_inter}} \hspace{0.5cm}
\subfloat[\centering]{\includegraphics[height=1.2in]{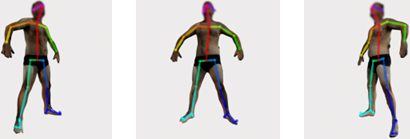}\label{fig:openpose}} 
\caption{(a) The avatar used as interface for translating points between different explorations. (b) Openpose detection (coloured lines) on three virtually rendered views of a patient’s forward-facing scan. }
\label{fig:avatar}
\end{figure*} 

\subsection{Anonymized 3D Avatar}
\label{section:avatar}

The 3D reconstruction process generates a detailed triangle mesh of the patient's skin surface, but this raw geometric data lacks understanding of human body structure and anatomy. To address this limitation, the system employs a deformable human model (DHM) that provides semantic understanding of body topology and enables more meaningful interaction with the scan data.

The DHM approach offers significant advantages for both single-scan analysis and longitudinal patient monitoring. Within a single examination, the DHM allows us to relate to semantic concepts and perform queries such as “show me all the moles on the left arm of the patient.” For patients undergoing multiple scans over time, the DHM serves as a standardized interface for comparing lesion locations across different examinations, enabling accurate tracking of changes in existing lesions and detection of new ones (see Fig. \ref{fig:avatar_inter}).

Additionally, it also provides full anonymization of the patient, since once fitted, the system can transform the patient-specific body shape into a generic avatar representation. This is useful in cases where the patient is uncomfortable seeing their naked body on the screen (a generic pose/shape instance of the DHM can be shown instead) as it preserves anonymity of patient data.

After a review of the state of the art (see \citep{Hasler09, Yang14, Pishchulin17, Anguelov23}), we decided to use the Skinned Multi-Person Linear Model (SMPL) \citep{Loper23}. The SMPL model is a realistic 3D representation of the human body based on skinning and blend shapes, learned from thousands of 3D body scans. It consists of a template mesh, which defines the 3D topology of the body using a fixed set of vertices and triangles, and a skeleton (i.e., a set of joints) associated with this mesh. This mesh and skeleton deform according to the following parameters:

\begin{itemize}
    \item \emph{Shape parameters}, summarizing the displacements (expansion or shrinkage) of the vertices of the template mesh according to some PCA-derived principal deformations. The model has $300$ shape parameters, although the first $10$ are enough to represent the main variations among human bodies (height, weight, waist size, ...). 
    
    \item \emph{Pose parameters} vector of size $24 \times 3$, containing $24$ joints with a 3D vector per-joint encoding a rotation in axis-angle representation that defines the relative rotation of each joint in the skeleton’s base shape. 
\end{itemize}

In order to achieve the mapping presented in Fig. \ref{fig:avatar_inter}, we must fit the DHM parameters to render the avatar shape so that it mimics the 3D surface mesh of the patient as closely as possible. The large number of parameters of the SMPL model makes direct optimization of all parameters against the 3D model impractical. Therefore, we divide the fitting process into two sequential steps.

First, we estimate the pose of the patient, that is, the 3D pose of the joints of the model’s skeleton. Although we have 3D data, recent solutions addressing the human pose estimation problem typically use 2D images \citep{Chen18, Sun19, Cao21}. Therefore, we apply the OpenPose \citep{Cao21} (see Fig. \ref{fig:openpose}) pose estimator to various 2D views of the reconstructed model, rendered using a virtual camera in slightly different positions. With multiple views of the 3D skeleton joints and the known positions of the virtual cameras, we can reconstruct those joints in 3D. 

Then, we optimize \citep{Kingma15} the parameters to minimize:

\begin{itemize}
    \item Discrepancies between the skeleton joint locations estimated using OpenPose and those resulting from the SMPL model’s pose and shape parameters. 
    \item  The Hausdorff distance between the actual 3D mesh of the patient’s body and the surface mesh of the instantiated SMPL model. 
\end{itemize}

A gender-specific SMPL model is used according to the patient, and the optimizer is constrained by a human pose prior \citep{Pavlakos19} that penalizes anatomically impossible joint rotations.

\subsection{Interface for the Operator}
\label{section:gui}

\begin{figure*}[t]
\centering
\subfloat[\centering]{\includegraphics[height=2.4in]{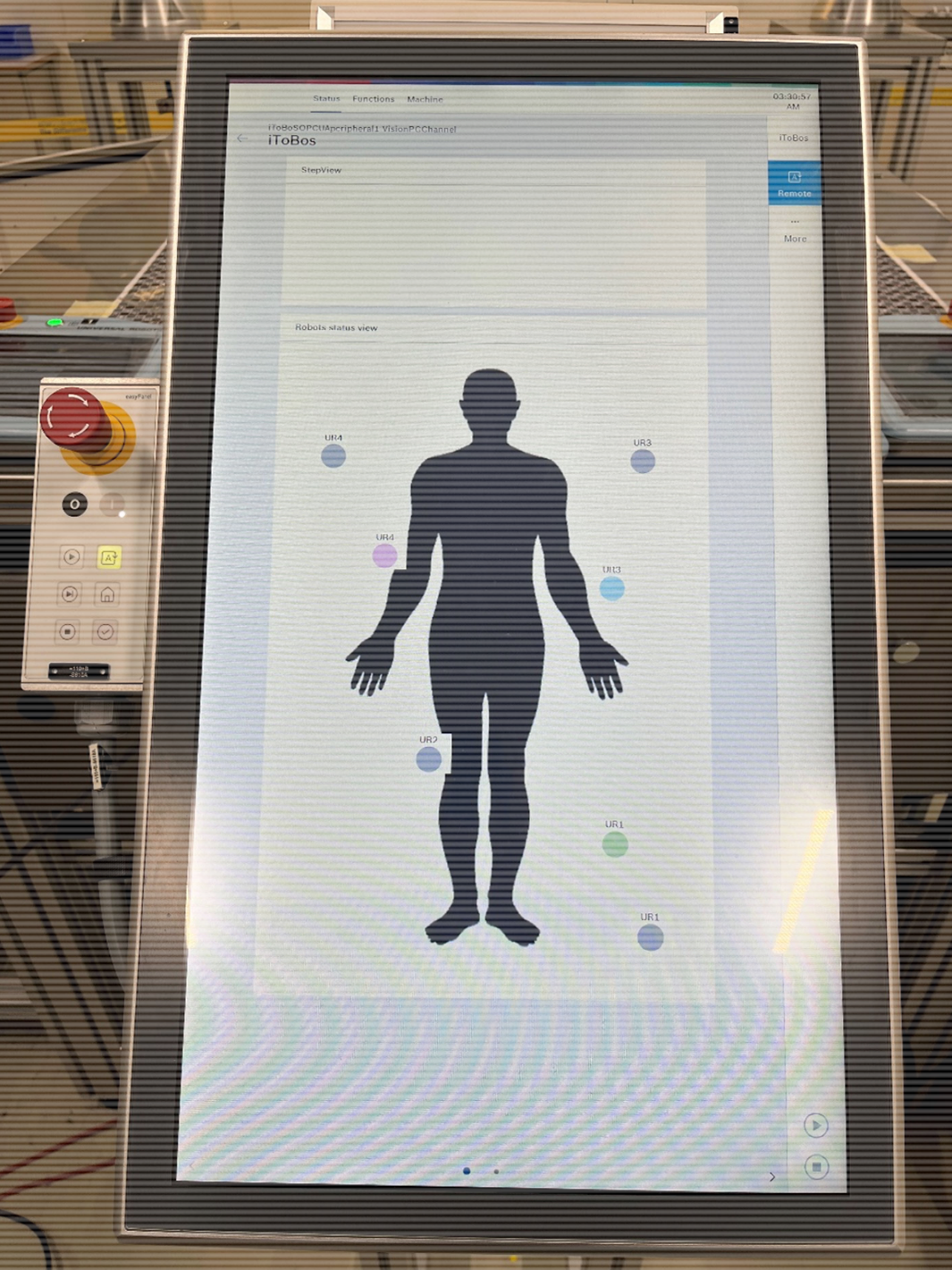}\label{fig:plc_touchscreen}} \hspace{0.5cm}
\subfloat[\centering]{\includegraphics[height=2.4in]{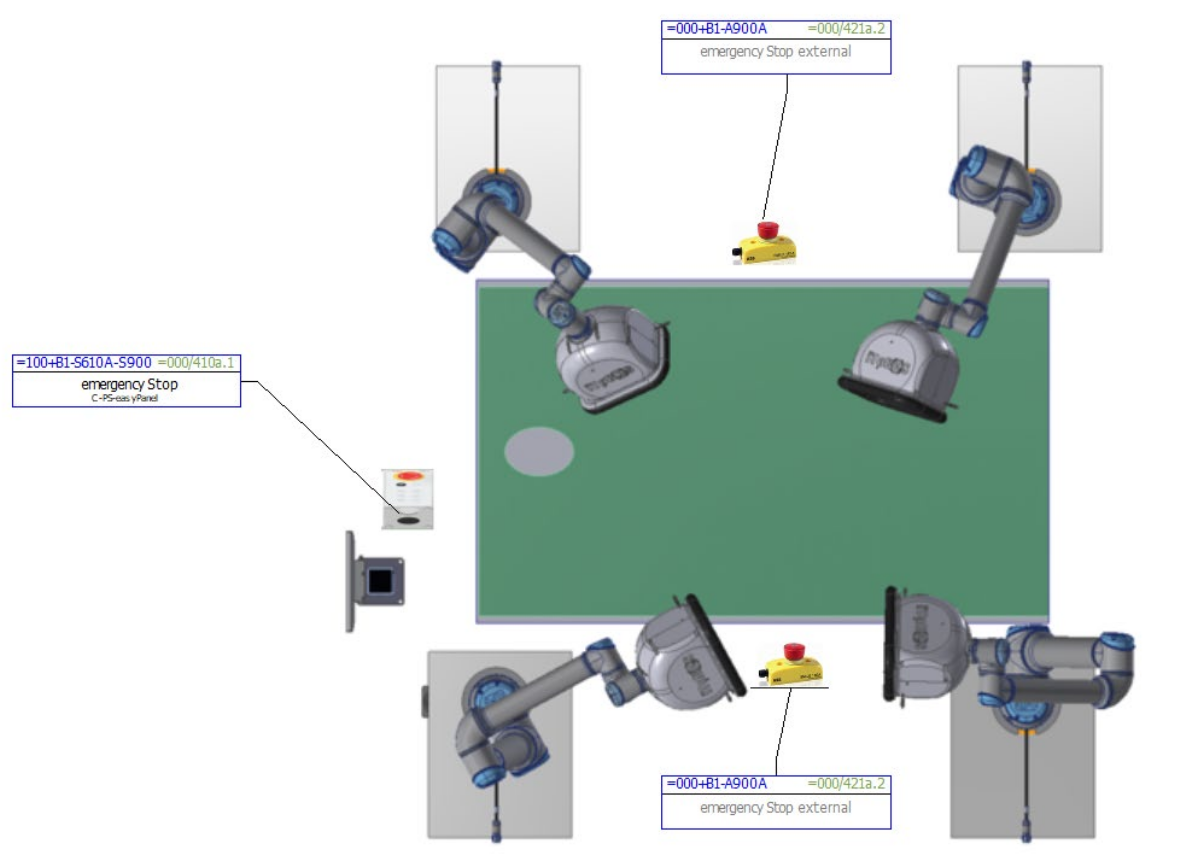}\label{fig:emergency}} 
\caption{(a) Touch screen display panel for the human operator, with the Bosch Easy Panel shown on the left. (b) Locations of the emergency stop buttons around the scanner.}
\label{fig:itobos_emergency}
\end{figure*} 

\subsubsection{Main control panel of scanner}

A PLC computer controls the mechatronic system of the scanner and hosts the vision system (described in Section \ref{section:gui} above). A $24$” touchscreen display is used for the human-machine interface (HMI). The HMI is connected to the computer vision system, allowing the operator to control it. On the right side of the display, the operator can choose the operational mode of the scanner.

\begin{itemize}
    \item Local: Used to activate manual functions and maintenance.
    \item Remote: The system operates as a \textit{child}, with the computer vision system acting as the \textit{parent}.
\end{itemize}

When \textit{Remote} mode is selected, the step view window guides the operator through the procedure for scanning a new patient. First, the system prompts the operator to enter the patient ID (which can be typed directly on the touchscreen) and then requests the operator to position the patient inside the machine. After the patient is settled and ready to be scanned, the operator should press the \textit{OK} button on the screen to start the full-body scan procedure (described in Section \ref{section:scanning_procedure}). 

\subsubsection{Physical interfaces}

 Several physical buttons on the panel are implemented to activate safety functions. These physical interfaces are integrated into a small panel (Bosch Easy Panel), an ergonomic device (Enable button), and the emergency stop buttons.

The \textit{Bosch Easy Panel } contains safety controls and shortcut buttons for the most common functions of the machine. This small control panel is placed next to the mattress, giving the operator one-click access to the following functions:

\begin{itemize}
    \item Emergency stop (1): all actuators are immediately halted, and the brakes are engaged. To rearm the machine, the emergency stop button must be turned and released.
    \item Control ON (2): powers the actuators of the machine.
    \item Control OFF (3): switches off all actuators of the scanner.
    \item Auto mode (4): switches to Auto mode (the standard scanning mode).
    \item Homing mode (5): moves the scanner to its initial position to accommodate a new patient.
    \item Acknowledge error (6): acknowledges any station errors or warnings.
    \item Play (7): starts the selected mode (Automatic or Homing).
    \item Pause (8): pauses the current process. To resume, the operator must press the play button again.
    \item Stop button (9): stops the scanning process. 
\end{itemize}

\subsubsection{Emergency stops}

Several emergency stop buttons are located around the machine, allowing the operator to stop it safely. When an emergency stop is activated, all actuators are turned off and the brakes on all axes are engaged. Figure \ref{fig:emergency} shows the locations of the emergency stop buttons. 

\begin{table}[!t]
\caption{Image resolution in $\frac{\text{px}}{\text{mm}}$ and $\frac{\upmu \text{m}}{\text{px}}$ or each image modality used during the dataset collection step. Pixel resolution reflects sensor sampling density, derived from sensor and optical specifications. True optical resolution was measured empirically using a USAF 1951 resolution target.}
\centering
\begin{tabular}{|@{}c@{}|@{}c@{}|@{}c@{}|}\hline
    \begin{tabular}{c}
       \textbf{Modality} \\\hline 
       Dermoscopy \\\hline
       Vectra \\\hline
       Native \\\hline
       SR \\\hline
    \end{tabular} &
    \begin{tabular}{c|c}
    \multicolumn{2}{c}{\textbf{Pixel resolution}}\\\hline 
       $\sim$ \SI{187}{px \per\milli\meter}   & $\sim$ \SI{5}{\micro\meter\per px}\\\hline 
       $\sim$ \SI{10}{px \per\milli\meter}    & $\sim$ \SI{105}{\micro\meter\per px}\\ \hline
       $\sim$ \SI{68}{px \per\milli\meter} &  $\sim$ \SI{15}{\micro\meter\per px}\\\hline
       $\sim$ \SI{137}{px \per\milli\meter} & $\sim$ \SI{7}{\micro\meter\per px}\\\hline
    \end{tabular}
    &
    \begin{tabular}{c}
    \textbf{True optical resolution} \\\hline
       \SI{8.8}{\micro\meter\per px} \\\hline
       N/A \\\hline
       \SI{22.1}{\micro\meter\per px} \\\hline
       \SI{22.1}{\micro\meter\per px} \\\hline
    \end{tabular}
\end{tabular}
\label{table:results}
\end{table}

\section{Experiments and Results}
\label{section:results}

\subsection{Data Collection}

To evaluate the image acquisition performance of our scanner, we collected a dataset of lesion images. The dataset includes images captured using three different complementary imaging modalities: a commercial total-body photography system (VECTRA WB360)\citep{Vectra}, conventional contact dermoscopy \citep{DermoscopeD200} (Canfield D200EVO), and our liquid lens-based automated non-contact-dermoscopic imaging system. These modalities represent the current clinical workflow (TBP followed by manual dermoscopy) and the proposed automated acquisition workflow, enabling direct comparison of image quality and optical performance.

The images captured with our scanner were further processed using two distinct post-processing pipelines, designed according to dermatologist feedback. This dual approach allows for a quantitative comparison between the two pipelines, facilitating the selection of a standardized processing method. 

Both post-processing techniques involve an initial image sharpening step, followed by super-resolution and contrast enhancement (described in Section \ref{sec:image_enhancement}). Their primary distinction lies in the super-resolution method employed: the first pipeline utilizes a pre-trained GPEN (GAN Prior Embedded Network) model \citep{Yang2021GPEN}, while the second applies a self-trained model based on a Real-ESRGAN (Enhanced Super-Resolution GAN) \citep{RealEsgran} with SRVGGNetCompact architecture.

To distinguish between these processing approaches, we refer to the two scanner-derived image sets as Native and SR. Native denotes images processed with the GPEN-based pipeline, maintaining the scanner's native resolution (\SI{\approx 68}{px\per\milli\meter}; \SI{\approx 15}{\micro\meter\per px}; Table \ref{table:results}), while SR denotes images processed with the Real-ESRGAN (SRVGGNetCompact) super-resolution pipeline, achieving enhanced resolution (\SI{\approx 137}{px\per\milli\meter}; \SI{\approx 7}{\micro\meter\per px}). Throughout the analysis, we use the abbreviated terms Native and SR for clarity.

The collected dataset includes a total of $156$ lesions, each captured using four image modalities (Vectra, dermoscopy, Native, and SR). Lesion sizes range from \SI{0.7}{\milli\meter} to \SI{16}{\milli\meter}, with an average of \SI[separate-uncertainty-units=single]{3.2(1.9)}{\milli\meter}. Most lesions fall within the \SI{2.2}{\milli\meter} to \SI{3.7}{\milli\meter} range,  indicating most lesions are relatively small.

\begin{figure*}[!t]
\centering
\subfloat[]{\includegraphics[width=0.35\linewidth]{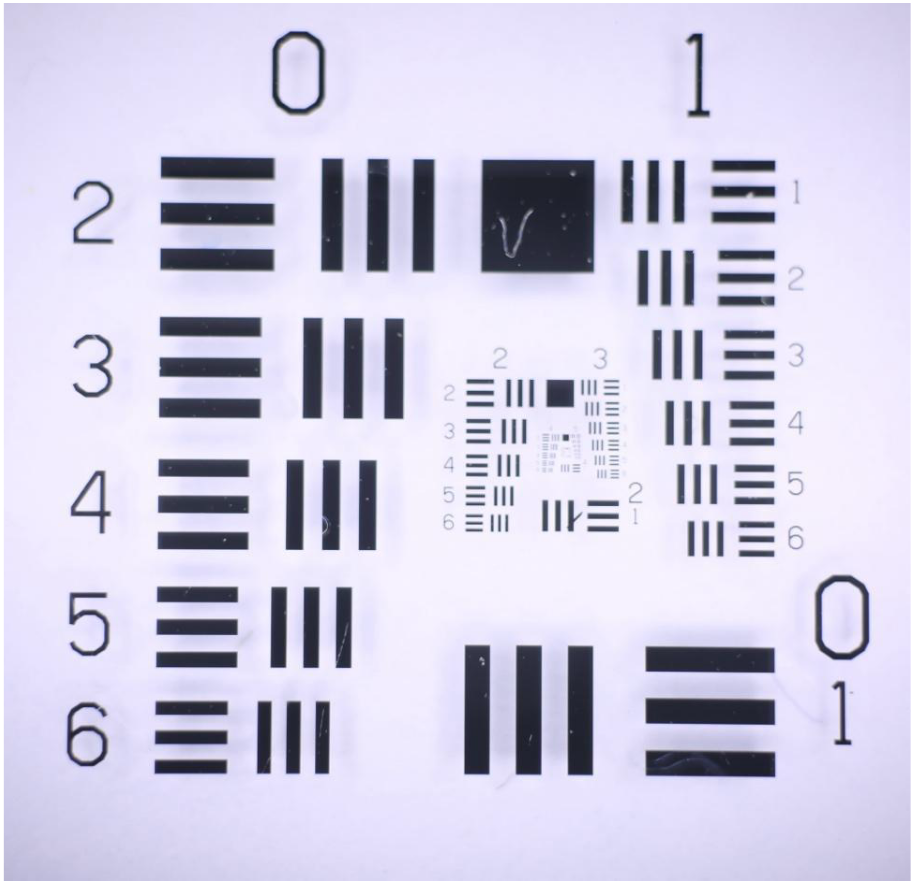}}
\hspace{1cm}
\subfloat[]{\includegraphics[width=0.35\linewidth]{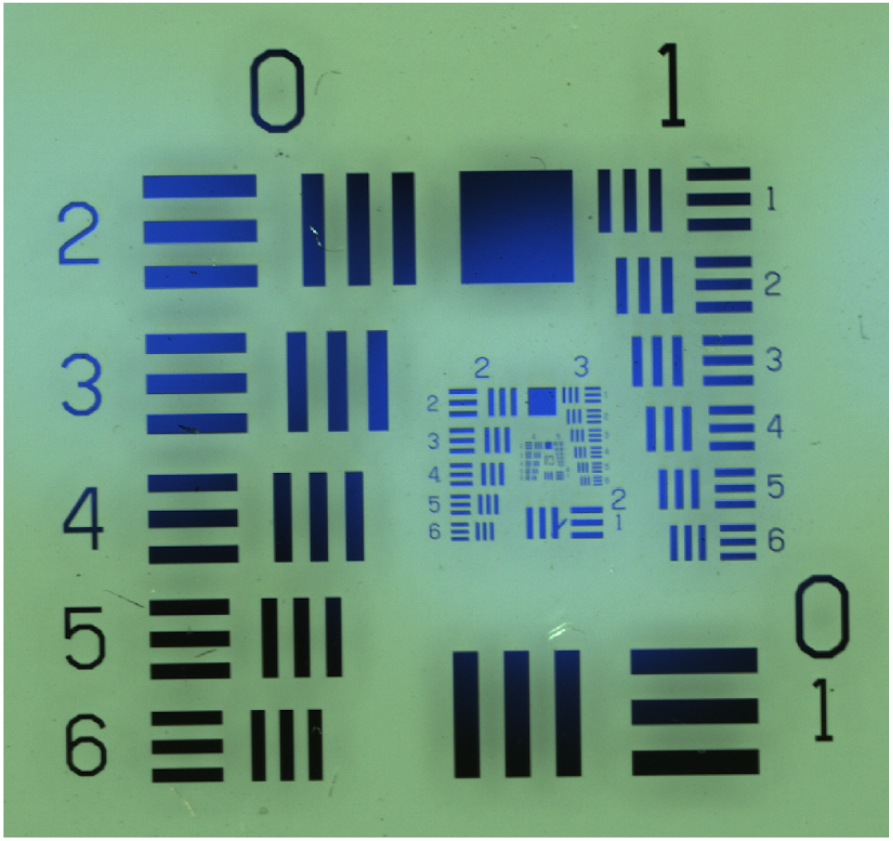}}
\caption{USAF 1951 resolution target imaged with (a) the contact dermatoscope, resolving Group 5, Element 6; and (b) our scanner at \SI{26}{\centi\meter}, resolving Group 4, Element 4.}
\label{fig:usaf_target}
\end{figure*}

Table \ref{table:results} provides an overview of the pixel resolution for each imaging modality, based on sensor specifications rather than a direct optical measurement. Resolution values for dermoscopy and our scanner (Native and SR) reflect this pixel resolution; the Vectra value is likewise an approximation based on the system's sensor specifications and has not been independently validated by the authors. It is important to distinguish this pixel resolution from the imaging system's true optical resolution, since image quality depends on the entire optical system, measured in line pairs per millimeter (lp/mm), rather than the megapixel count of the camera alone; true resolving power is determined by the combined quality of lenses, optical design, and sensor characteristics. To quantify this, we imaged a USAF 1951 resolution target with the contact dermatoscope and our scanner at their respective working distances (contact for the dermatoscope, \SI{26}{\centi\meter} for our scanner), shown in Figure \ref{fig:usaf_target} . The dermatoscope resolved Group 5, Element 6, while our scanner resolved Group 4, Element 4. These readings were converted to line pairs per millimeter and then to a smallest resolvable feature size using standard USAF 1951 conversion formulas \citep{EdmundOpticsUSAF} , yielding a true optical resolution of \SI{8.8}{\micro\meter} for the dermatoscope, compared to \SI{22.1}{\micro\meter} for our scanner, measured identically for both the Native and SR pipelines. This indicates that, despite increasing pixel sampling density from \SI{\approx 68}{px\per\milli\meter} to \SI{\approx 137}{px\per\milli\meter}, the super-resolution post-processing pipeline improves apparent image detail without increasing the system's true optical resolving power.

\sisetup{uncertainty-mode = separate}

\subsection{Comparison of feature quality across imaging modalities}

Image quality is paramount in dermoscopy, as high-resolution and well-lit images are essential for accurate analysis. Poor image quality can obscure clinically relevant features, limiting the usefulness of the resulting images for lesion assessment. Optimal dermoscopic images should have adequate magnification (usually $10$x or greater), proper lighting, and minimal distortion, ensuring that features are clearly visible. Key structures to assess during dermoscopy include pigment network, globules, streaks, regression structures, and vascular patterns. 

%Some of these features (e.g., regression areas, blue-white veil, shiny white streaks) were observed in only a small number of lesions in our dataset, reflecting their natural rarity in the patient population studied rather than a limitation of any imaging modality's ability to visualize them when present.

\begin{figure*}[t]
\centering

\subfloat{\includegraphics[width=1.6in, angle=90]{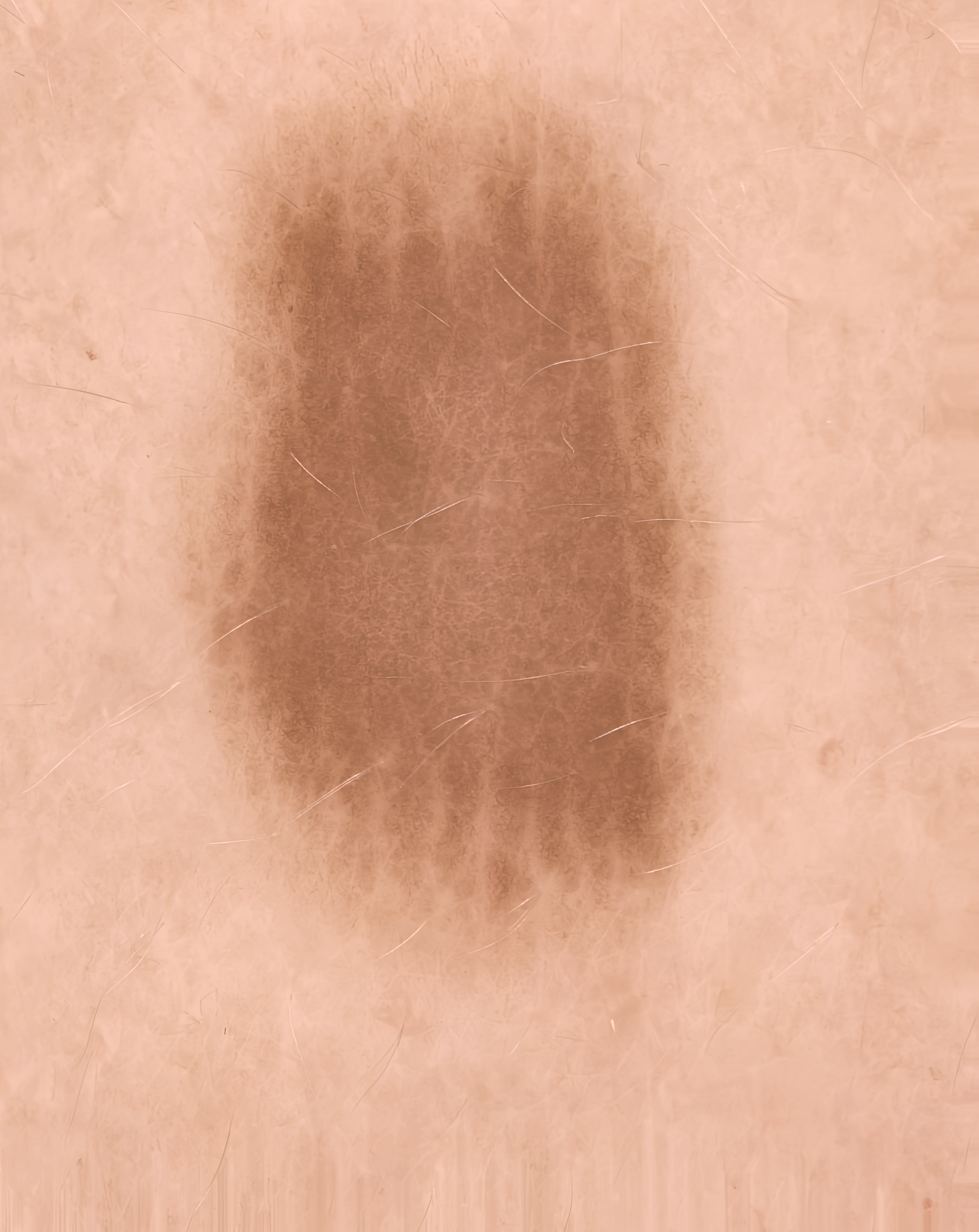}} \hspace{0.5cm}
\subfloat{\includegraphics[width=1.6in, angle=90]{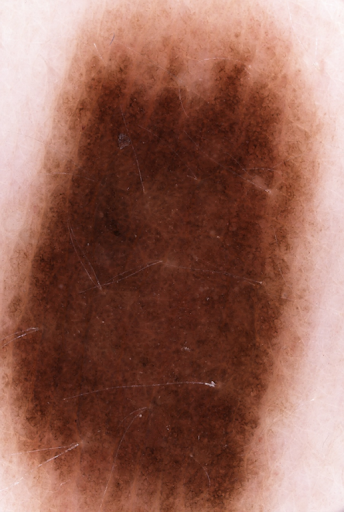}} \hspace{0.5cm}
\subfloat{\includegraphics[width=1.6in, angle=90]{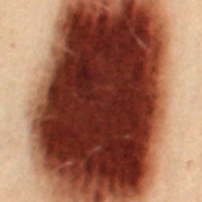}}

\vspace{0.5cm}

\subfloat{\includegraphics[height=1.6in]{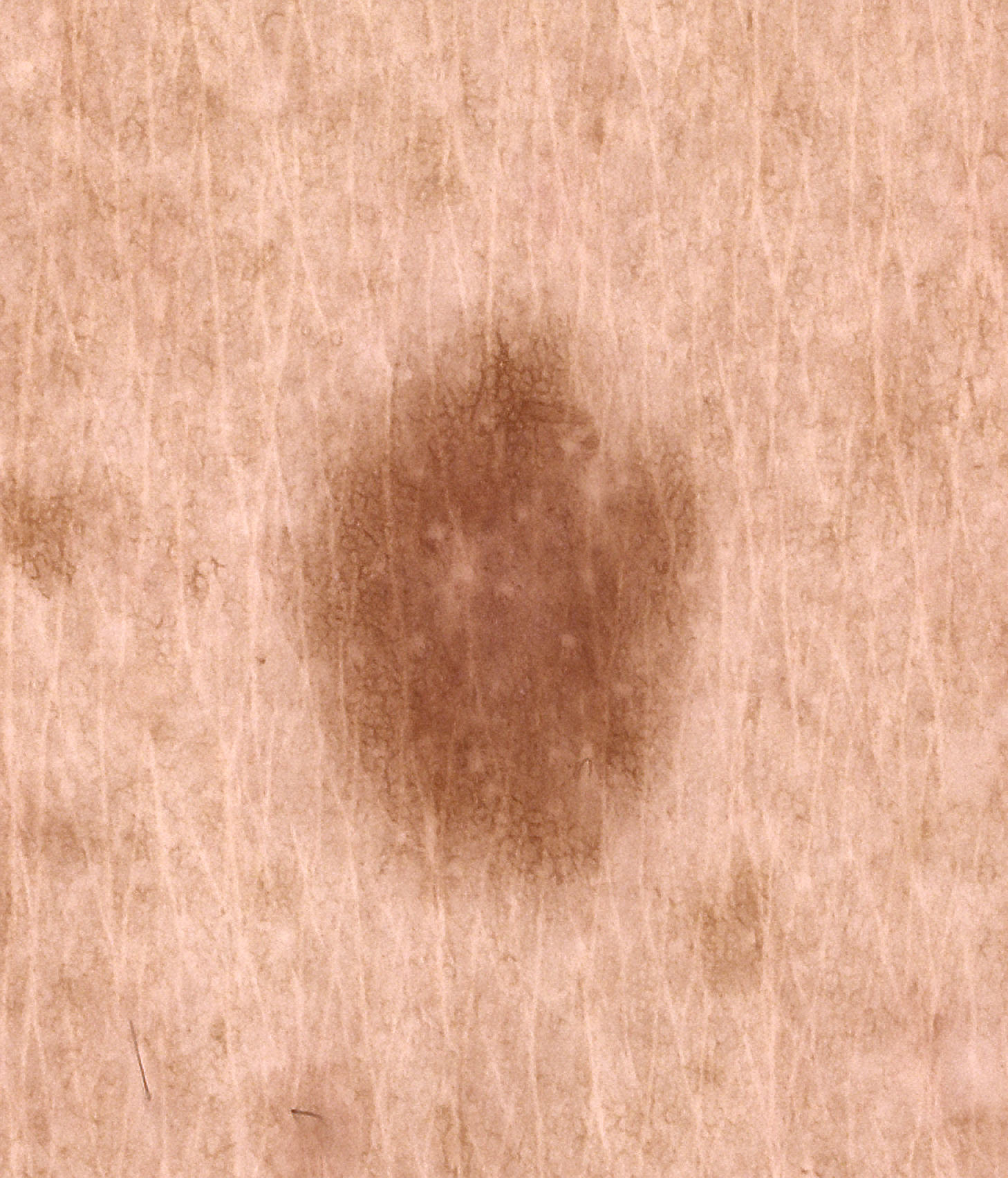}} \hspace{0.5cm}
\subfloat{\includegraphics[height=1.6in]{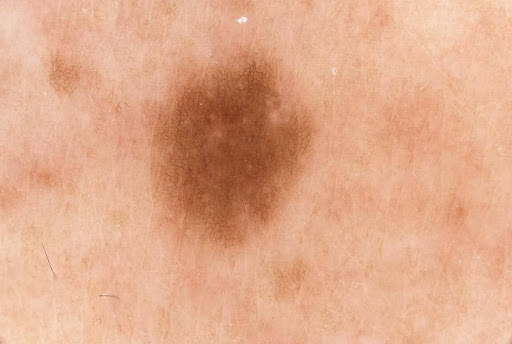}} \hspace{0.5cm}
\subfloat{\includegraphics[height=1.6in]{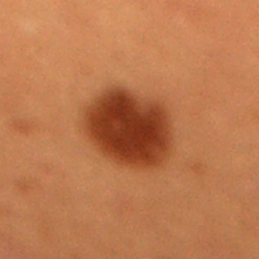}}
\caption{Comparison of dermoscopic images obtained with the scanner, hand-held dermoscope and Vectra TBP. (a) and (d) Native, (b) and (e) hand-held dermoscope and (c) and (f) Vectra TBP.}
\label{fig:itobos_dermoscopy}
\end{figure*}

\begin{figure*}[t]
\centering
\subfloat[\centering]{\includegraphics[width=\textwidth]{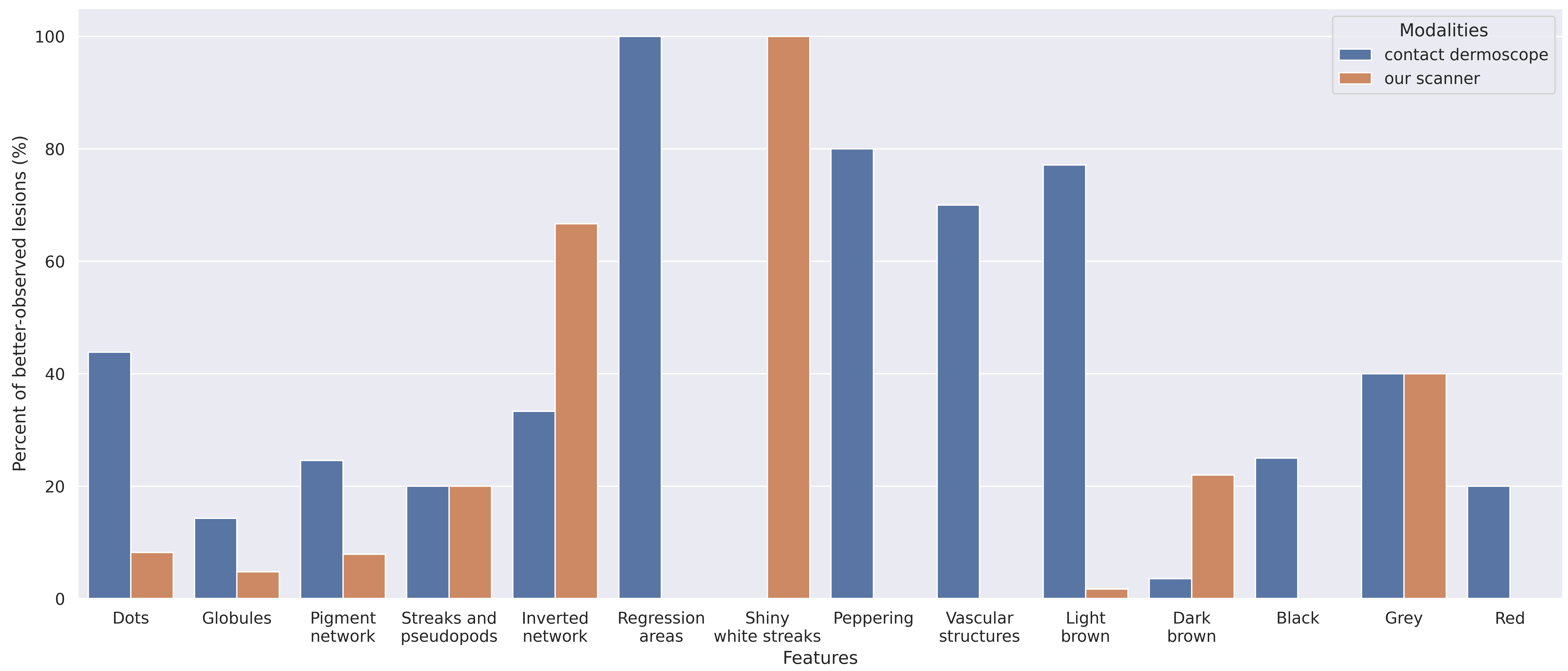}\label{fig:dermo_vs_scanner}} \hspace{0.5cm}
\subfloat[\centering]{\includegraphics[width=\textwidth]{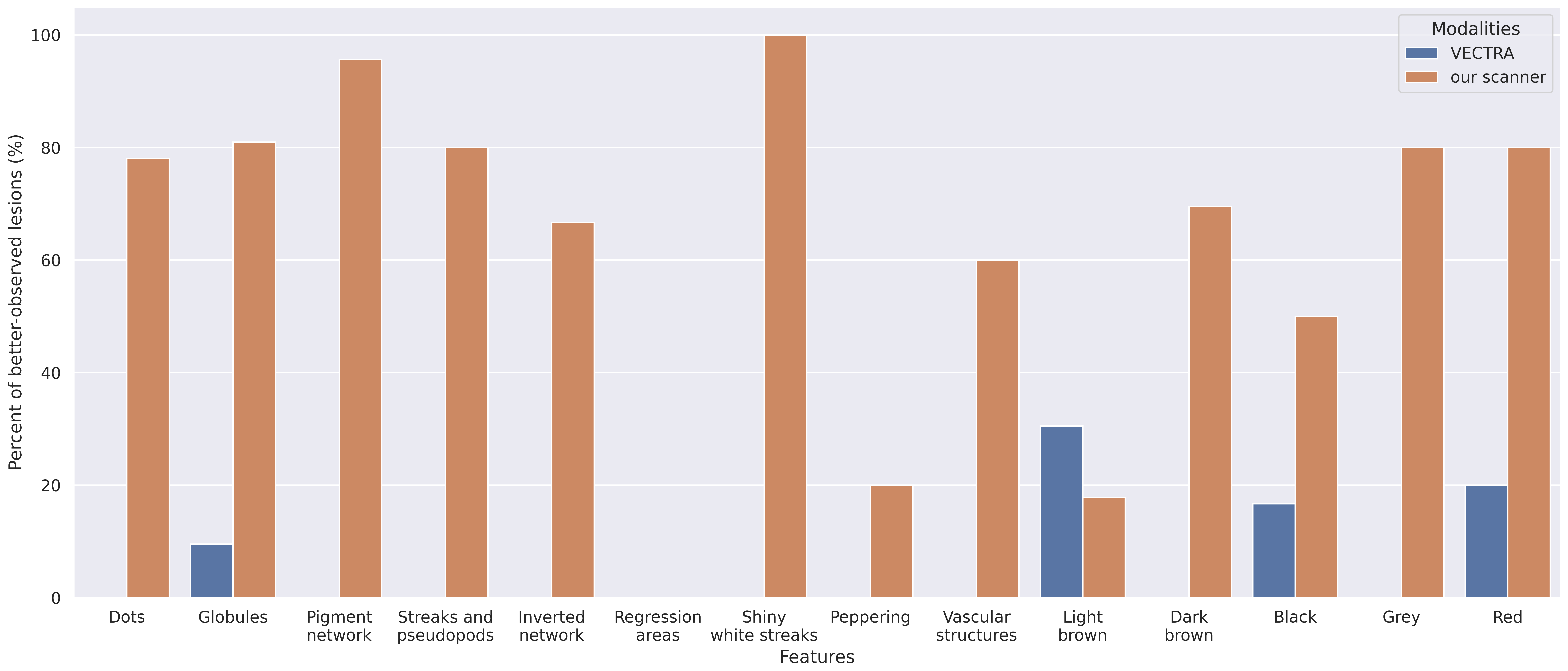}\label{fig:vectra_vs_scanner}} 
\caption{Comparative analysis of lesion feature visualization quality across three imaging modalities. It quantifies the percentage of the cases in which each modality outperforms the other. (a) Shows a comparison between contact dermoscopy and our scanner, and (b) between VECTRA and our scanner.}
\label{fig:image_quality_comparisons}
\end{figure*}

The acquired lesion images have undergone detailed evaluation by two expert dermatologists, who assessed 17 clinically relevant features, categorizing them as \textit{excellent}, \textit{good}, \textit{mild}, or \textit{not visible}. These features included structural characteristics such as dots, globules, pigmented network, streaks and pseudopods, inverted network, regression areas, blue-white veil, shiny white streaks, peppering and vascular structures. Additionally, pigmentation colours --- including light and dark brown, black, blue, white, grey and red --- were also assessed. The pigment network, characterized by a regular pattern of dark lines and light areas, is typically seen in benign lesions such as nevi. Irregularities in this pattern may raise suspicion for malignancy. Globules, which are small, round clusters of pigment, are common in benign lesions but can be irregular or large in melanoma. Streaks, particularly in atypical nevi or melanoma, can indicate an aggressive lesion. Regression structures, such as white or grey areas with a scar-like appearance, may suggest a malignant process. Vascular patterns, like dotted or linear vessels, are also significant in distinguishing benign from malignant lesions, with more irregular vascular patterns being indicative of malignancy. The three columns in Fig. \ref{fig:itobos_dermoscopy} display images captured from the scanner (Native), contact dermoscope and Vectra, respectively.

When comparing contact dermoscopy to the scanner modalities, Native and SR exhibit a closer performance (Fig. \ref{fig:dermo_vs_scanner}), with dermoscopy surpassing them in certain features but showing competitive results in others. Dermoscopy proves superior in capturing regression areas, as neither of the scanner images is able to capture this feature. This modality also shows better performance at capturing peppering, where \SI{80}{\percent} of cases favour dermoscopy while \SI{20}{\percent} show equal performance, vascular structures (\SI{60}{\percent} of cases in favour of dermoscopy, \SI{40}{\percent} equal), and light brown coloration (\SI{77}{\percent} in favour of dermoscopy, \SI{21}{\percent} equal). For the remaining pigmentation properties (dark brown, black, grey and red) both Native and SR show comparable performance, suggesting that scanner images provide a viable alternative for colour assessment. 

On the other hand, scanner images outperform contact dermoscopy in capturing shiny white streaks, with \SI{100}{\percent} of cases favouring our scanner. The visualization of the inverted network is also better captured by our scanner, with \SI{67}{\percent} of cases showing improved visualization and \SI{33}{\percent} favouring dermoscopy. For the remaining features analysed, including globules, pigment network, and streaks and pseudopods, both modalities exhibit comparable performance, highlighting the potential of our scanner as an alternative acquisition method to contact dermoscopy for these features.

This level of feature visibility is achieved despite the gap in true optical resolving power reported earlier in this section (\SI{22.1}{\micro\meter} for our scanner versus \SI{8.8}{\micro\meter} for contact dermoscopy). Even though super-resolution improves pixel sampling density without improving true resolving power, most evaluated lesion features were still captured at comparable quality by our scanner, with our scanner outperforming for shiny white streaks and inverted network, and dermoscopy remaining superior for regression areas, peppering, vascular structures and light-brown pigmentation.

The comparison between Vectra and the scanner modalities highlights the limitations of Vectra in visualising essential lesion structures (Fig. \ref{fig:vectra_vs_scanner}). Across most features, scanner images consistently outperform Vectra, particularly in the capture of dots (\SI{78}{\percent} of cases favour our scanner vs. \SI{0}{\percent} favour Vectra), globules (\SI{81}{\percent} vs. \SI{10}{\percent}), pigment network (\SI{96}{\percent} vs. \SI{0}{\percent}), streaks and pseudopods (\SI{80}{\percent} vs. \SI{0}{\percent}), inverted network (\SI{67}{\percent} vs. \SI{0}{\percent}) and vascular structures (\SI{60}{\percent} vs. \SI{0}{\percent}). Furthermore, pigmentation features are also better captured in the scanner images, including dark brown (\SI{70}{\percent} vs. \SI{0}{\percent}), black (\SI{50}{\percent} vs. \SI{17}{\percent}), grey (\SI{80}{\percent} vs. \SI{0}{\percent}) and red (\SI{80}{\percent} vs. \SI{20}{\percent}), reinforcing the superior capability of our scanner for colour differentiation.

However, in light brown colouration, Vectra demonstrates slightly better performance relative to Native and SR, with \SI{31}{\percent} of cases favouring Vectra compared to \SI{18}{\percent} for our scanner. Additionally, the visualization of peppering appears to be comparable between the two modalities. These findings indicate that while Vectra may offer some advantages in specific pigmentation properties, its overall ability to capture key lesion structures remains limited compared to our scanner.

\begin{figure}[t]
\centering
\subfloat{\includegraphics[width=0.45\textwidth]{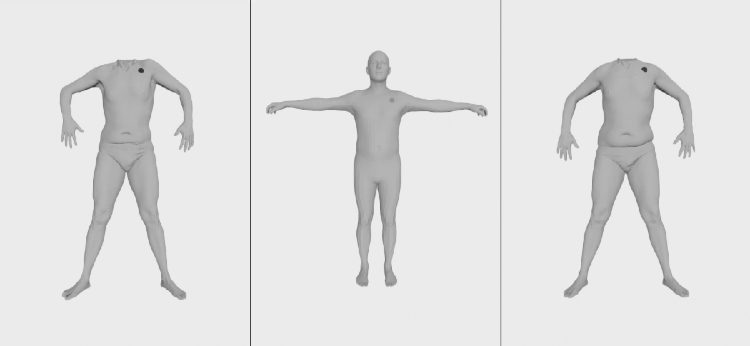}} \hspace{0.4cm}
\subfloat{\includegraphics[width=0.45\textwidth]{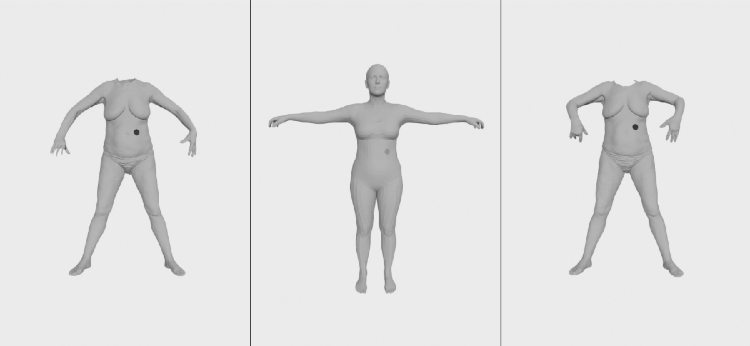}}

\vspace{0.5cm}

\subfloat{\includegraphics[width=0.45\textwidth]{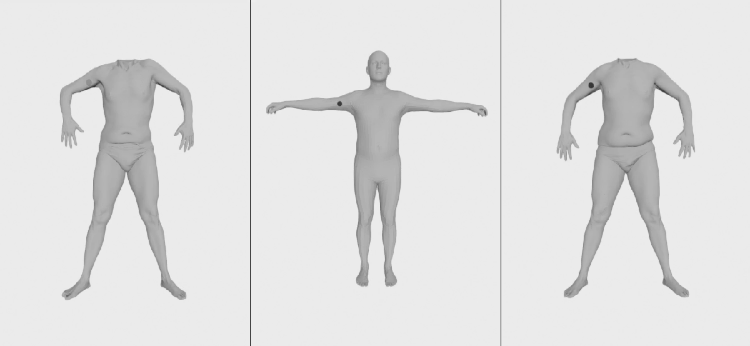}} \hspace{0.4cm}
\subfloat{\includegraphics[width=0.45\textwidth]{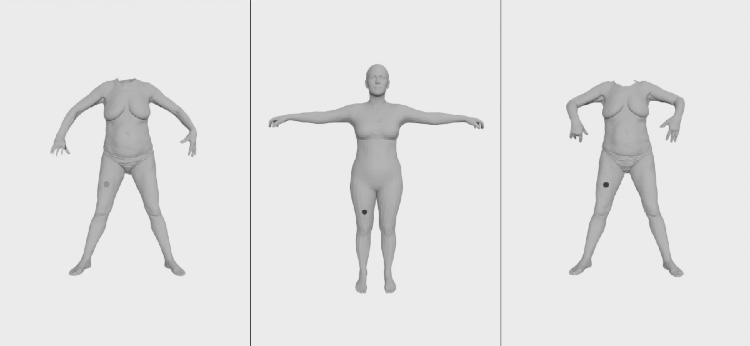}}
\caption{Using the avatar as an interface between two explorations. The mesh of one exploration is shown in the right side, the mean avatar (a generic instantiation of the SMPL model, not related to the meshes at all) is on the middle, and on the left side there is a mesh of another exploration. (a) and (c) show examples with a male body model, while (b) and (d) examples with a female body model. In (a) and (b), a point in the avatar (green) is translated to the two explorations (red). In the same way, the green point selected in the first exploration's model in (c) and (d) is translated to both the avatar and the second exploration.}
\label{fig:avatar_transfer}
\end{figure}

These findings indicate that while Vectra may offer some advantages in specific pigmentation properties, its overall ability to capture key lesion structures remains limited compared to our scanner. Part of this limitation stems from Vectra's fixed camera geometry: because the 92 cameras are mounted at fixed positions and distances, the effective resolution achieved for a given lesion depends on that patient's body shape and position relative to the fixed camera array, and can vary substantially between patients. In contrast, our scanner's view-planning system (Section \ref{sec:view_planning}) actively repositions each cobot's camera relative to the reconstructed 3D surface of the individual patient, maintaining a consistent working distance and viewing angle for each lesion regardless of body shape or size. This adaptive positioning is expected to provide more consistent image quality across a diverse patient population, compared to a fixed-camera system where resolution at any given lesion depends on variable patient geometry.

This variability in effective resolution across patients is consistent with existing reports on high-resolution total body photography (HR-TBP). As documented by Primiero et al. \citep{Primiero2024} , such systems offer substantially lower image resolution than dermoscopy. Similarly, Baete et al. \citep{Baete2025} conclude in their review that despite being suitable for lesion change detection, dermoscopy remains necessary for diagnostic-level assessment. This matches how such systems are described by Ferrera et al. \citep{NonContactFerrera} : as producing \textit{pseudodermoscopic} images rather than \textit{true dermoscopic} images. These are high-resolution, polarized-light captures that approximate the look of dermoscopy and support lesion localization and general pattern recognition, but do not resolve fine structures such as pigment network, dots/globules, and vascular patterns at dermatoscopic-level detail. Our feature-level comparison against Vectra corroborates this distinction empirically.

\subsection{Avatar Creation}

\begin{figure*}[!t]
\centering
\subfloat{\includegraphics[height=1.25in]{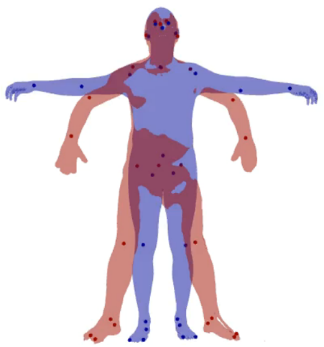}}
\hspace{1cm}
\subfloat{\includegraphics[height=1.25in]{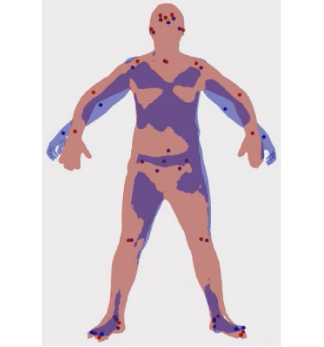}}
\hspace{1cm}
\subfloat{\includegraphics[height=1.25in]{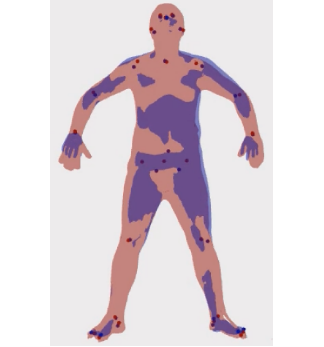}}
\hspace{1cm}
\subfloat{\includegraphics[height=1.25in]{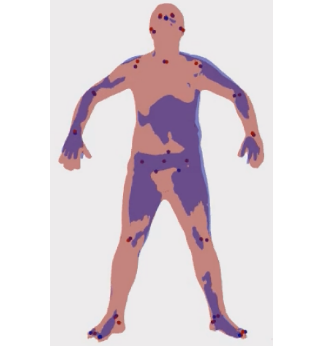}}
\caption{Different steps of the optimization process, showing at the same time the reference measurements (in red) and the SMPL model being optimized (in blue). Starting from (a) the optimization changes the shape and pose parameters of the SMPL model to fit both the surfaces (depicted as transparent meshes) and the joints' location (depicted as small spheres). After passing through the intermediate steps (b) and (c) note in (d) how the final fitting converges to a SMPL model that closely matches the shape and pose of the patient in the scan.}
\label{fig:avatar_optim}
\end{figure*}

In order to validate the optimization process behind the avatar creation, we inspect the proposed optimization procedure under non-ideal initializations. Fig. \ref{fig:avatar_optim} shows that, even when initialized away from the true pose of the patient, the optimization process is able to properly fit the pose parameters first and then focus on the shape parameters. Some intermediate steps of the optimization can be seen in Fig. \ref{fig:avatar_optim}. Obviously, initializing the optimization with the pose estimation provided by OpenPose greatly speeds up convergence.

We also validate the use of the avatar as an interface between explorations by transferring some positions from one of the meshes obtained in one exploration to another via the DHM representation. Fig. \ref{fig:avatar_transfer} shows several examples of these transfers for both male and female patients, demonstrating these consistent spatial correspondence.

\section{Conclusions and Future Work}
\label{section:conclusion}

We presented a full-body skin-imaging system composed of four collaborative UR10 manipulators, each carrying 2D, 3D and dermoscopic (liquid-lens) cameras arranged around a medical bed to acquire images at near-dermoscopic quality. The scanner is complemented by software for 3D body reconstruction, automatic mole detection, view planning, camera calibration and an avatar-based masking interface that preserves patient anonymity.

In our study of $156$ lesions imaged with four modalities (Vectra, contact dermoscopy, and two scanner pipelines), we report the scanner’s native sampling of \SI{\approx 68}{px \per \milli \meter} (\SI{\approx 15}{\micro\meter \per px}) and its super-resolved variant at \SI{137}{px \per \milli \meter} (\SI{\approx 7}{\micro\meter \per px}). To avoid confusion between hardware and post-processing, we refer to the two scanner pipelines as Native (GPEN-based enhancement at native sampling) and SR (Real-ESRGAN/SRVGGNetCompact super-resolution). This dual-pipeline design enables a quantitative comparison to inform a standardized processing choice.

Across clinically relevant features, the scanner images consistently outperformed Vectra (notably for dots, globules, pigment network, streaks/pseudopods, inverted network and vascular structures) while Vectra showed an advantage only for light-brown pigmentation and peppering was broadly comparable. Relative to contact dermoscopy, Native and SR achieved competitive quality for most clinical features, with dermoscopy remaining superior for regression areas, peppering, vascular structures and light-brown pigmentation, despite a measured gap in true optical resolving power relative to contact dermoscopy. These results support the potential of our scanner as a non-contact dermoscopy alternative for many features while retaining a safe, contact-free workflow. In addition, we validated the anonymized avatar interface as a practical bridge across explorations by transferring locations between scans using the DHM representation, demonstrating consistent spatial correspondence on male and female examples. 

The demonstrated image quality improvements over existing total body photography systems show that automated full-body dermoscopic image acquisition is technically feasible while providing substantially richer lesion detail than commercial TBP systems.  Current practice requires dermatologists to manually examine each suspicious lesion with handheld dermoscopes after initial TBP screening, resulting in two separate image acquisition steps. Our system integrates these steps into a single automated acquisition workflow by capturing dermoscopic-quality images during the full-body examination, reducing the need for separate manual dermoscopic image acquisition.

The contact-free nature of our approach offers additional advantages for infection control and patient comfort, particularly relevant in post-pandemic healthcare environments and for patients with sensitive or compromised skin conditions. The automated 3D mapping eliminate the need for manual positioning of the dermoscope over individual lesions, enabling a fully-automated dermoscopic image acquisition process. Although this automation is expected to simplify the imaging workflow, quantitative comparisons of acquisition time clinical workflow efficiency remain subjects for future work.

Moreover, the avatar-based anonymization system addresses growing concerns about medical data privacy while enabling privacy-preserving longitudinal comparison of patient examinations. By providing a consistent anatomical reference across repeated scans, it facilitates the localization and monitoring of pigmented skin lesions over time while preserving patient anonymity.

This paper demonstrates the technical feasibility of automated, full-body dermoscopic imaging at clinically relevant quality standards. While significant validation work remains before clinical deployment, the proposed system demonstrates that robotic automation can integrate total-body photography and dermoscopic image acquisition into a single imaging workflow , reducing reliance on separate manual dermoscopic examinations. Future work will focus on larger-scale clinical validation, wokflow optimization, and further improvements in image quality and acquisition speed. The path to clinical adoption will require continued collaboration between engineering, dermatology and regulatory communities to ensure that technical capabilities translate into improved patient outcomes, but the results presented here establish a foundation for future robotic systems capable of automated, high-quality dermoscopic image acquisition in clinical practice.

\section*{Acknowledgements}

This research was funded by the Spanish government through the project ASSiST (PID2023-149413OB-I00).

% Use the style of numbering in square brackets.
% If nothing is used, default style will be taken.
%\begin{enumerate}[a)]
%\item 
%\item 
%\item 
%\end{enumerate}  

%\begin{itemize}
%\item 
%\item 
%\item 
%\end{itemize}  

%\begin{description}
%\item[]
%\item[] 
%\item[] 
%\end{description}  

\clearpage

\printcredits

\bibliographystyle{cas-model2-names}

\bibliography{cas-refs}

@article{FryerDistortion,
author = {Fryer, John and Brown, Duane},
year = {1986},
month = {01},
pages = {51-58},
title = {Lens distortion for close-range photogrammetry},
volume = {52},
journal = {Photogrammetric Engineering and Remote Sensing}
}

@article{JuradoAruco,
author = {Garrido-Jurado, Sergio and Muñoz-Salinas, Rafael and Madrid-Cuevas, Francisco and Marín-Jiménez, Manuel},
year = {2014},
month = {06},
pages = {2280–2292},
title = {Automatic generation and detection of highly reliable fiducial markers under occlusion},
volume = {47},
journal = {Pattern Recognition},
doi = {10.1016/j.patcog.2014.01.005}
}

@article{Santos1997EvaluationOA,
  title={Evaluation of autofocus functions in molecular cytogenetic analysis},
  author={Andres Santos and C. Ortiz De Sol{\'o}rzano and J. J. Vaquero and Jos{\'e}-Mar{\'i}a Pe{\~n}a and Norberto Malpica and F. del Pozo},
  journal={Journal of Microscopy},
  year={1997},
  volume={188},
  url={https://api.semanticscholar.org/CorpusID:25729585}
}

@misc{UR10e,
  title        = {UR10e},
  author       = {Universal Robots},
  year         = 2024,
  note         = {\url{https://www.universal-robots.com/products/ur10-robot/}}
}

@misc{EvetarOptics,
  title        = {Evetar Leading Optics},
  author       = {Evetar},
  year         = 2024,
  note         = {\url{http://www.leadingoptics.com/}}
}

@misc{DermoscopeD200,
  title        = {D200evo},
  author       = {Canfield Scientific, Inc.},
  year         = 2025,
  note         = {\url{https://www.canfieldsci.com/imaging-systems/visiomed-d200evo/}}
}

@misc{Vectra,
  title        = {VECTRA WB360},
  author       = {Canfield Scientific, Inc.},
  year         = 2025,
  note         = {\url{https://www.medicalexpo.it/prod/canfield/product-124461-1103627.html}}
}

@Article{SkinPiotr,
AUTHOR = {Szczypiński, Piotr M. and Sprawka, Katarzyna},
TITLE = {Orthorectification of Skin Nevi Images by Means of 3D Model of the Human Body},
JOURNAL = {Sensors},
VOLUME = {21},
YEAR = {2021},
NUMBER = {24},
ARTICLE-NUMBER = {8367},
URL = {https://www.mdpi.com/1424-8220/21/24/8367},
PubMedID = {34960467},
ISSN = {1424-8220},
DOI = {10.3390/s21248367}
}

@article{TBPWinkler,
title = {Performance of an automated total body mapping algorithm to detect melanocytic lesions of clinical relevance},
journal = {European Journal of Cancer},
volume = {202},
pages = {114026},
year = {2024},
issn = {0959-8049},
doi = {https://doi.org/10.1016/j.ejca.2024.114026},
url = {https://www.sciencedirect.com/science/article/pii/S0959804924006828},
author = {Julia K. Winkler and Katharina S. Kommoss and Ferdinand Toberer and Alexander Enk and Lara V. Maul and Alexander A. Navarini and Jeremy Hudson and Gabriel Salerni and Albert Rosenberger and Holger A. Haenssle}
}

@misc{DermoScan,
  title        = {DermoScan X2},
  author       = {DermoScan GmbH},
  year         = 2025,
  note         = {\url{https://dermatoskop.sk/dermoscan-x2/}}
}

@article{Hasler09,
    author = {Hasler, N. and Stoll, C. and Sunkel, M. and Rosenhahn, B. and Seidel, H.-P.},
    title = {A Statistical Model of Human Pose and Body Shape},
    journal = {Computer Graphics Forum},
    volume = {28},
    number = {2},
    pages = {337-346},
    doi = {https://doi.org/10.1111/j.1467-8659.2009.01373.x},
    url = {https://onlinelibrary.wiley.com/doi/abs/10.1111/j.1467-8659.2009.01373.x},
    eprint = {https://onlinelibrary.wiley.com/doi/pdf/10.1111/j.1467-8659.2009.01373.x},
    year = {2009}
}

@inproceedings{Yang14,
  author={Yang, Yipin and Yu, Yao and Zhou, Yu and Du, Sidan and Davis, James and Yang, Ruigang},
  booktitle={2014 2nd International Conference on 3D Vision}, 
  title={Semantic Parametric Reshaping of Human Body Models}, 
  year={2014},
  volume={2},
  number={},
  pages={41-48},
  doi={10.1109/3DV.2014.47}
}

@article{Pishchulin17,
    author = {Pishchulin, Leonid and Wuhrer, Stefanie and Helten, Thomas and Theobalt, Christian and Schiele, Bernt},
    title = {Building statistical shape spaces for 3D human modeling},
    year = {2017},
    issue_date = {July 2017},
    publisher = {Elsevier Science Inc.},
    address = {USA},
    volume = {67},
    number = {C},
    issn = {0031-3203},
    url = {https://doi.org/10.1016/j.patcog.2017.02.018},
    doi = {10.1016/j.patcog.2017.02.018},
    journal = {Pattern Recogn.},
    month = jul,
    pages = {276–286},
    numpages = {11}
}

@inbook{Anguelov23,
author = {Anguelov, Dragomir and Srinivasan, Praveen and Koller, Daphne and Thrun, Sebastian and Rodgers, Jim and Davis, James},
title = {SCAPE: Shape Completion and Animation of People},
year = {2023},
isbn = {9798400708978},
publisher = {Association for Computing Machinery},
address = {New York, NY, USA},
edition = {1},
url = {https://doi.org/10.1145/3596711.3596797},
booktitle = {Seminal Graphics Papers: Pushing the Boundaries, Volume 2},
articleno = {85},
numpages = {9}
}

@article{Loper23,
    author = {Loper, Matthew and Mahmood, Naureen and Romero, Javier and Pons-Moll, Gerard and Black, Michael J.},
    title = {SMPL: a skinned multi-person linear model},
    year = {2015},
    issue_date = {November 2015},
    publisher = {Association for Computing Machinery},
    address = {New York, NY, USA},
    volume = {34},
    number = {6},
    issn = {0730-0301},
    url = {https://doi.org/10.1145/2816795.2818013},
    doi = {10.1145/2816795.2818013},
    journal = {ACM Trans. Graph.},
    month = oct,
    articleno = {248},
    numpages = {16}
}

@article{Chen18,
    Author = {Chen, Yilun and Wang, Zhicheng and Peng, Yuxiang and Zhang, Zhiqiang and Yu, Gang and Sun, Jian},
    Title = {{Cascaded Pyramid Network for Multi-Person Pose Estimation}},
    Conference = {CVPR},
    Year = {2018}
}

@inproceedings{Sun19,
  title={Deep High-Resolution Representation Learning for Human Pose Estimation},
  author={Sun, Ke and Xiao, Bin and Liu, Dong and Wang, Jingdong},
  booktitle={CVPR},
  year={2019}
}

@article{Cao21,
    author={Cao, Zhe and Hidalgo, Gines and Simon, Tomas and Wei, Shih-En and Sheikh, Yaser},
    journal={ IEEE Transactions on Pattern Analysis \& Machine Intelligence },
    title={{ OpenPose: Realtime Multi-Person 2D Pose Estimation Using Part Affinity Fields }},
    year={2021},
    volume={43},
    number={01},
    ISSN={1939-3539},
    pages={172-186},
    doi={10.1109/TPAMI.2019.2929257},
    url = {https://doi.ieeecomputersociety.org/10.1109/TPAMI.2019.2929257},
    publisher={IEEE Computer Society},
    address={Los Alamitos, CA, USA},
    month=jan
}

@inproceedings{Pavlakos19,
  title = {Expressive Body Capture: 3D Hands, Face, and Body from a Single Image},
  author = {Pavlakos, Georgios and Choutas, Vasileios and Ghorbani, Nima and Bolkart, Timo and Osman, Ahmed A. A. and Tzionas, Dimitrios and Black, Michael J.},
  booktitle = {Proceedings IEEE Conf. on Computer Vision and Pattern Recognition (CVPR)},
  year = {2019}
}

@inproceedings{Kingma15,
  author = {Kingma, Diederik P. and Ba, Jimmy},
  booktitle = {ICLR (Poster)},
  editor = {Bengio, Yoshua and LeCun, Yann},
  ee = {http://arxiv.org/abs/1412.6980},
  title = {Adam: A Method for Stochastic Optimization.},
  year = 2015
}

@INPROCEEDINGS{YoloVarghese,
  author={Varghese, Rejin and M., Sambath},
  booktitle={2024 International Conference on Advances in Data Engineering and Intelligent Computing Systems (ADICS)}, 
  title={YOLOv8: A Novel Object Detection Algorithm with Enhanced Performance and Robustness}, 
  year={2024},
  volume={},
  number={},
  pages={1-6},
  doi={10.1109/ADICS58448.2024.10533619}}

@inproceedings{Yang2021GPEN,
    title={{GAN} Prior Embedded Network for Blind Face Restoration in the Wild},
    author={Yang, Tao and Ren, Peiran and Xie, Xuansong and Zhang, Lei},
    booktitle={IEEE Conference on Computer Vision and Pattern Recognition (CVPR)},
    year={2021}
}

@INPROCEEDINGS{RealEsgran,
  author={Wang, Xintao and Xie, Liangbin and Dong, Chao and Shan, Ying},
  booktitle={2021 IEEE/CVF International Conference on Computer Vision Workshops (ICCVW)}, 
  title={Real-ESRGAN: Training Real-World Blind Super-Resolution with Pure Synthetic Data}, 
  year={2021},
  volume={},
  number={},
  pages={1905-1914},
  doi={10.1109/ICCVW54120.2021.00217}}

@misc{Lumo,
  title        = {LumoScanner System},
  author       = {Lumo Imaging},
  year         = 2025,
  note         = {\url{https://lumoscan.com/dermatology/lumoscanner-system/}}
}

@misc{ATBM,
  title        = {ATBM master 4th Generation},
  author       = {FotoFinder Systems GmbH},
  year         = 2025,
  note         = {\url{https://www.fotofinder.de/en/technology/total-body-dermoscopy/bodystudio-atbm/master}}
}

@article{ValidationVectra,
author = {De Stefani, Alberto and Barone, Martina and Alamdari, Sam and Barjami, Arjola and Baciliero, Ugo and Apolloni, Federico and Gracco, Antonio and Bruno, Giovanni},
year = {2022},
month = {07},
pages = {8820},
title = {Validation of Vectra 3D Imaging Systems: A Review},
volume = {19},
journal = {International Journal of Environmental Research and Public Health},
doi = {10.3390/ijerph19148820}
}

@article{MelanomaVectra,
author = {Rayner, Jenna and Laino, Antonia and Nufer, Kaitlin and Adams, Laura and Raphael, Anthony and Menzies, Scott and Soyer, Peter},
year = {2018},
month = {05},
pages = {},
title = {Clinical Perspective of 3D Total Body Photography for Early Detection and Screening of Melanoma},
volume = {5},
journal = {Frontiers in Medicine},
doi = {10.3389/fmed.2018.00152}
}

@article{AustraliaVectra,
author="Horsham, Caitlin
and O'Hara, Montana
and Sanjida, Saira
and Ma, Samantha
and Jayasinghe, Dilki
and Green, Adele C
and Schaider, Helmut
and Aitken, Joanne F
and Sturm, Richard A
and Prow, Tarl
and Soyer, H Peter
and Janda, Monika",
title="The Experience of 3D Total-Body Photography to Monitor Nevi: Results From an Australian General Population-Based Cohort Study",
journal="JMIR Dermatol",
year="2022",
month="Jun",
day="20",
volume="5",
number="2",
pages="e37034",
issn="2562-0959",
doi="10.2196/37034",
url="https://doi.org/10.2196/37034"
}

@article{EfficacyVectra,
title = {Comparison of the efficacy of skin examination using 3D total body photography to clinical and dermoscopic examination},
journal = {EJC Skin Cancer},
volume = {2},
pages = {100264},
year = {2024},
issn = {2772-6118},
doi = {https://doi.org/10.1016/j.ejcskn.2024.100264},
url = {https://www.sciencedirect.com/science/article/pii/S2772611824002520},
author = {Frank Friedrich Gellrich and Anne Strunk and Julian Steininger and Friedegund Meier and Stefan Beissert and Sarah Hobelsberger}
}

@misc{VectraNews,
  title        = {News and Updates},
  author       = {Canfield Scientific, Inc.},
  year         = 2025,
  note         = {\url{https://www.canfieldsci.com/in-the-news/}}
}

@ARTICLE{EvangelidisImageAlignment,
  author={Evangelidis, Georgios D. and Psarakis, Emmanouil Z.},
  journal={IEEE Transactions on Pattern Analysis and Machine Intelligence}, 
  title={Parametric Image Alignment Using Enhanced Correlation Coefficient Maximization}, 
  year={2008},
  volume={30},
  number={10},
  pages={1858-1865},
  doi={10.1109/TPAMI.2008.113}}

@article{KanjarImageSharpness,
title = {Image Sharpness Measure for Blurred Images in Frequency Domain},
journal = {Procedia Engineering},
volume = {64},
pages = {149-158},
year = {2013},
note = {International Conference on Design and Manufacturing (IConDM2013)},
issn = {1877-7058},
doi = {https://doi.org/10.1016/j.proeng.2013.09.086},
url = {https://www.sciencedirect.com/science/article/pii/S1877705813016007},
author = {Kanjar De and V. Masilamani}
}

@INPROCEEDINGS{SongImageFusion,
  author={Song, Le and Lin, Yuchi and Feng, Weichang and Zhao, Meirong},
  booktitle={2009 International Workshop on Intelligent Systems and Applications}, 
  title={A Novel Automatic Weighted Image Fusion Algorithm}, 
  year={2009},
  volume={},
  number={},
  pages={1-4},
  doi={10.1109/IWISA.2009.5072656}}

@INPROCEEDINGS{BuadesImageDenoising,
  author={Buades, A. and Coll, B. and Morel, J.-M.},
  booktitle={2005 IEEE Computer Society Conference on Computer Vision and Pattern Recognition (CVPR'05)}, 
  title={A non-local algorithm for image denoising}, 
  year={2005},
  volume={2},
  number={},
  pages={60-65 vol. 2},
  doi={10.1109/CVPR.2005.38}}

@Article{ViewPlanningFranchi,
AUTHOR = {Franchi, Valerio and Campos, Ricard and Quintana, Josep and Gracias, Nuno and Garcia, Rafael},
TITLE = {Safety-Enforcing and Occlusion-Aware Camera View Planning for Full-Body Imaging},
JOURNAL = {Technologies},
VOLUME = {14},
YEAR = {2026},
NUMBER = {4},
ARTICLE-NUMBER = {197},
URL = {https://www.mdpi.com/2227-7080/14/4/197},
ISSN = {2227-7080},
DOI = {10.3390/technologies14040197}
}

@article{NonContactFerrera,
author = {Ferrera, Nuria and Caño, Abel and Campos, Ricard and Carrera, Cristina and Garcia, Rafael and González-Villà, Sandra and Hayat, Hassan and Lenoir, Clément and Nazari, Sana and Podlipnik, Sebastian and Puig, Susana and Quintana, Josep and Ricart, Narcis and Roth, Bernhard and Saha, Anup and Serra, Laura and Ventura, Mark and Malvehy, Josep},
title = {Long Distance Non-Contact Dermoscopy: Technological Foundations, Clinical Applications, and Future Directions},
journal = {International Journal of Dermatology},
volume = {n/a},
number = {n/a},
pages = {},
doi = {https://doi.org/10.1111/ijd.70560},
url = {https://onlinelibrary.wiley.com/doi/abs/10.1111/ijd.70560},
eprint = {https://onlinelibrary.wiley.com/doi/pdf/10.1111/ijd.70560}
}

@misc{EdmundOpticsUSAF,
  author       = {{Edmund Optics}},
  title        = {1951 {USAF} Resolution Calculator},
  howpublished = {\url{https://www.edmundoptics.com/knowledge-center/tech-tools/1951-usaf-resolution}},
  note         = {Accessed: 2026-07-24},
  year         = {2025}
}

@ARTICLE{Primiero2024,
    
AUTHOR={Primiero, Clare A.  and Betz-Stablein, Brigid  and Ascott, Nathan  and D’Alessandro, Brian  and Gaborit, Seraphin  and Fricker, Paul  and Goldsteen, Abigail  and González-Villà, Sandra  and Lee, Katie  and Nazari, Sana  and Nguyen, Hang  and Ntouskos, Valsamis  and Pahde, Frederik  and Pataki, Balázs E.  and Quintana, Josep  and Puig, Susana  and Rezze, Gisele G.  and Garcia, Rafael  and Soyer, H. Peter  and Malvehy, Josep },
           
TITLE={A protocol for annotation of total body photography for machine learning to analyze skin phenotype and lesion classification},
          
JOURNAL={Frontiers in Medicine},
          
VOLUME={Volume 11 - 2024},
  
YEAR={2024},
  
URL={https://www.frontiersin.org/journals/medicine/articles/10.3389/fmed.2024.1380984},
  
DOI={10.3389/fmed.2024.1380984},
  
ISSN={2296-858X}}

@Article{Baete2025,
author="Baete, Fran
and Jakers, Alyssa Laura
and Delanoye, Emilie
and Vande Velde, Nele
and Voet, Griet",
title="3D Total Body Photography as a Promising Innovation for Early Skin Cancer Detection: Scoping Review",
journal="JMIR Dermatol",
year="2025",
month="Dec",
day="17",
volume="8",
pages="e68510",
issn="2562-0959",
doi="10.2196/68510",
url="https://derma.jmir.org/2025/1/e68510",
url="https://doi.org/10.2196/68510"
}



\end{document}